\documentclass[10pt,twocolumn,letterpaper]{article}

\usepackage{iccv}
\usepackage{times}
\usepackage{epsfig}
\usepackage{graphicx}
\usepackage{amsmath}
\usepackage{amssymb}
\usepackage{booktabs}
\usepackage{tabu}

\usepackage{multirow}
\usepackage[table]{xcolor}
\definecolor{maroon}{cmyk}{0,0.87,0.68,0.32}

\usepackage{mathtools}
\usepackage{pifont}
\newcommand{\xmark}{\ding{55}}%

\DeclareMathOperator*{\argmaxB}{argmax}

\usepackage[pagebackref=true,breaklinks=true,letterpaper=true,colorlinks,bookmarks=false]{hyperref}

\iccvfinalcopy 

\def\iccvPaperID{10811} 

\ificcvfinal\fi

\begin{document}

\title{MARS: Model-agnostic Biased Object Removal without Additional Supervision for Weakly-Supervised Semantic Segmentation}


\author{
    Sanghyun Jo$^{1}$, In-Jae Yu$^{2}$, and Kyungsu Kim$^{3}$\thanks{Correspondence to}\\ 
    {$^{1}$OGQ, Seoul, Korea} \qquad {$^{2}$Samsung Electronics, Suwon, Korea}\\
    {$^{3}$Department of Data Convergence and Future Medicine, Sungkyunkwan University, Seoul, Korea}\\ 
  \texttt{\{shjo.april, ijyu.phd, kskim.doc\}@gmail.com}\\
} 

\maketitle
\ificcvfinal\thispagestyle{empty}\fi

\begin{abstract}
\vspace{-0.3cm}
Weakly-supervised semantic segmentation aims to reduce labeling costs by training semantic segmentation models using weak supervision, such as image-level class labels. However, most approaches struggle to produce accurate localization maps and suffer from false predictions in class-related backgrounds (\emph{i.e.}, biased objects), such as detecting a railroad with the train class. Recent methods that remove biased objects require additional supervision for manually identifying biased objects for each problematic class and collecting their datasets by reviewing predictions, limiting their applicability to the real-world dataset with multiple labels and complex relationships for biasing. Following the first observation that biased features can be separated and eliminated by matching biased objects with backgrounds in the same dataset, we propose a fully-automatic/model-agnostic biased removal framework called MARS (\textbf{M}odel-\textbf{A}gnostic biased object \textbf{R}emoval without additional \textbf{S}upervision), which utilizes semantically consistent features of an unsupervised technique to eliminate biased objects in pseudo labels. Surprisingly, we show that MARS achieves new state-of-the-art results on two popular benchmarks, PASCAL VOC 2012 (val: 77.7\%, test: 77.2\%) and MS COCO 2014 (val: 49.4\%), by consistently improving the performance of various WSSS models by at least 30\% without additional supervision. Code is available at \url{https://github.com/shjo-april/MARS}. 
   
\end{abstract}

\vspace{-0.5cm}

\section{Introduction}

Fully-supervised semantic segmentation (FSSS) \cite{chen2017deeplab, chen2018encoder}, which aims to classify each pixel of an image, requires time-consuming tasks and significant domain expertise in some applications \cite{yu2018methods} to prepare pixel-wise annotations. By contrast, weakly-supervised semantic segmentation (WSSS) with image-level supervision, which is the most economical among weak supervision, such as bounding boxes \cite{dai2015boxsup}, scribbles \cite{lin2016scribblesup}, and points \cite{bearman2016s}, reduces the labeling cost by more than $20\times$ \cite{bearman2016s}. The multi-stage learning framework is the dominant approach for training WSSS models with image-level labels. Since this framework heavily relies on the quality of initial class activation maps (CAMs), numerous researchers \cite{ahn2018learning, wang2020self, lee2021anti, chen2022class, xie2022clims, jo2022recurseed} moderate the well-known drawback of CAMs, highlighting the most discriminative part of an object to reduce the false negative (FN).

\begin{figure}[t]
  \centering
  \includegraphics[width=0.90\linewidth]{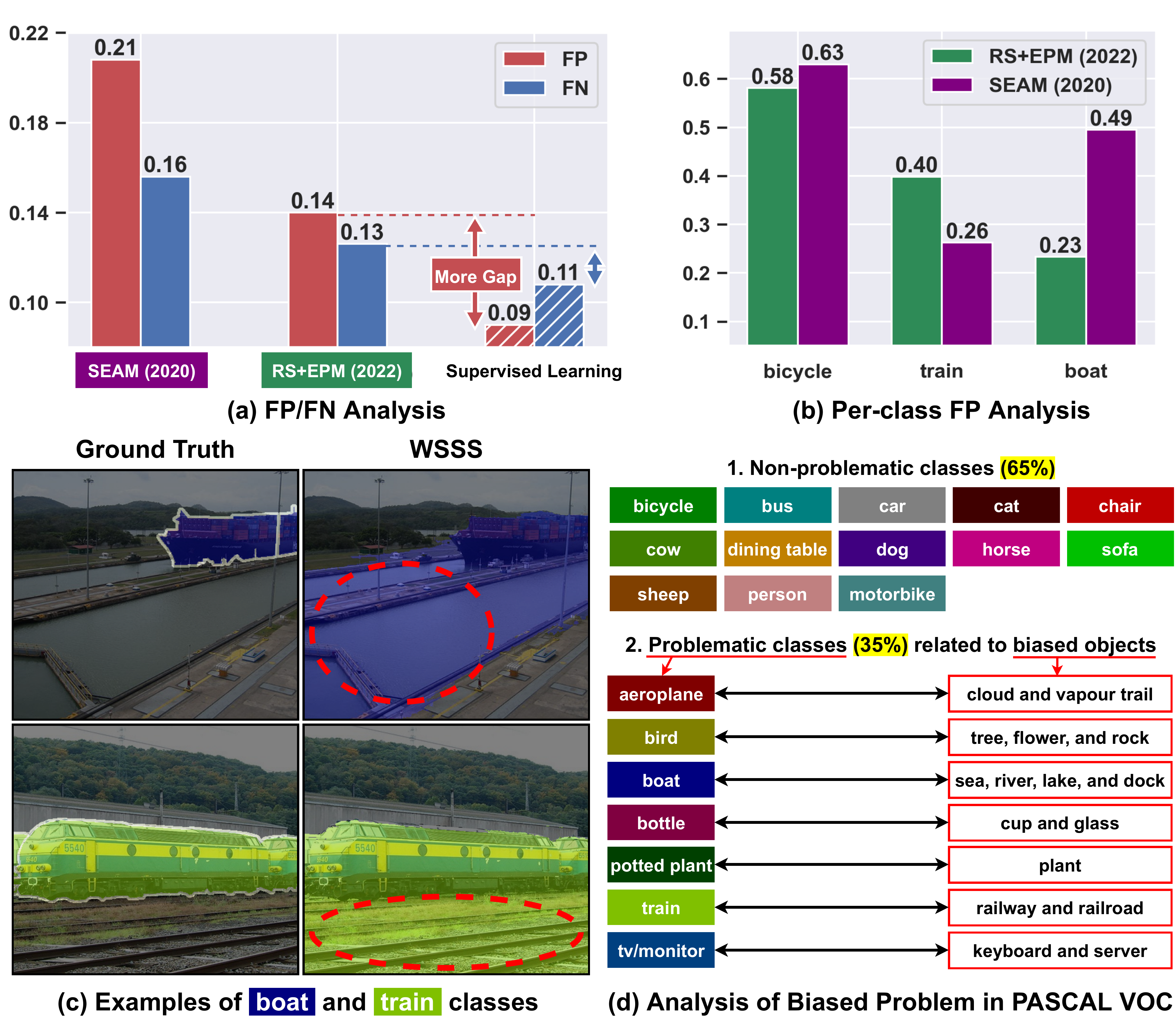}
  \caption{
      (a) Comparison with existing WSSS studies \cite{wang2020self, jo2022recurseed} and FSSS. (b) Per-class FP analysis. (c) Examples of biased objects in boat and train classes. (d) Quantitative analysis of biased objects on the PASCAL VOC 2012 dataset. Red dotted circles illustrate the false activation of biased objects such as railroad and sea.
  }
  \label{fig:problem}
  \vspace{-0.5cm}
\end{figure}

However, the false positive (FP) is the most crucial bottleneck to narrow the performance gap between WSSS and FSSS in Fig. \ref{fig:problem}(a). According to per-class FP analysis in Fig. \ref{fig:problem}(b), predicting target classes (\emph{e.g.}, boat) with class-related objects (\emph{e.g.}, sea) are factored into increasing FP in Fig. \ref{fig:problem}(c), besides incorrect annotations in the bicycle class. Moreover, 35\% of classes in the PASCAL VOC 2012 dataset have biased objects in Fig. \ref{fig:problem}(d). These results show that the performance degradation of previous approaches depends on the presence or absence of problematic classes in the dataset. We call this issue a biased problem. We also add examples of all classes in the Appendix.

\begin{table}
  \caption{  
    Comparison with public datasets for WSSS. Since Open Images \cite{kuznetsova2020open} does not provide pixel-wise annotations for all classes, existing methods employ PASCAL VOC 2012 \cite{everingham2010pascal} and MS COCO 2014 \cite{lin2014microsoft} for fair comparison and evaluation.
  } 
  \centering
  \vspace{+0.1cm}
  \begin{scriptsize} 
  \begin{tabular}{p{0.22\textwidth} | c c c}
    \toprule
    Dataset & Training images & Classes & GT \\  
    \hline 
    PASCAL VOC 2012 \cite{everingham2010pascal} & 10,582 & 20 & \checkmark \\
    MS COCO 2014 \cite{lin2014microsoft} & 80,783 & 80 & \checkmark \\
    Open Images \cite{kuznetsova2020open} & 9,011,219 & 19,794 & \xmark \\
    \bottomrule
  \end{tabular}
  \label{tab:dataset}
  \end{scriptsize}
  \vspace{-0.2cm}
\end{table}

\begin{figure}[t]
  \centering
  \includegraphics[width=0.9\linewidth, height=4.5cm]{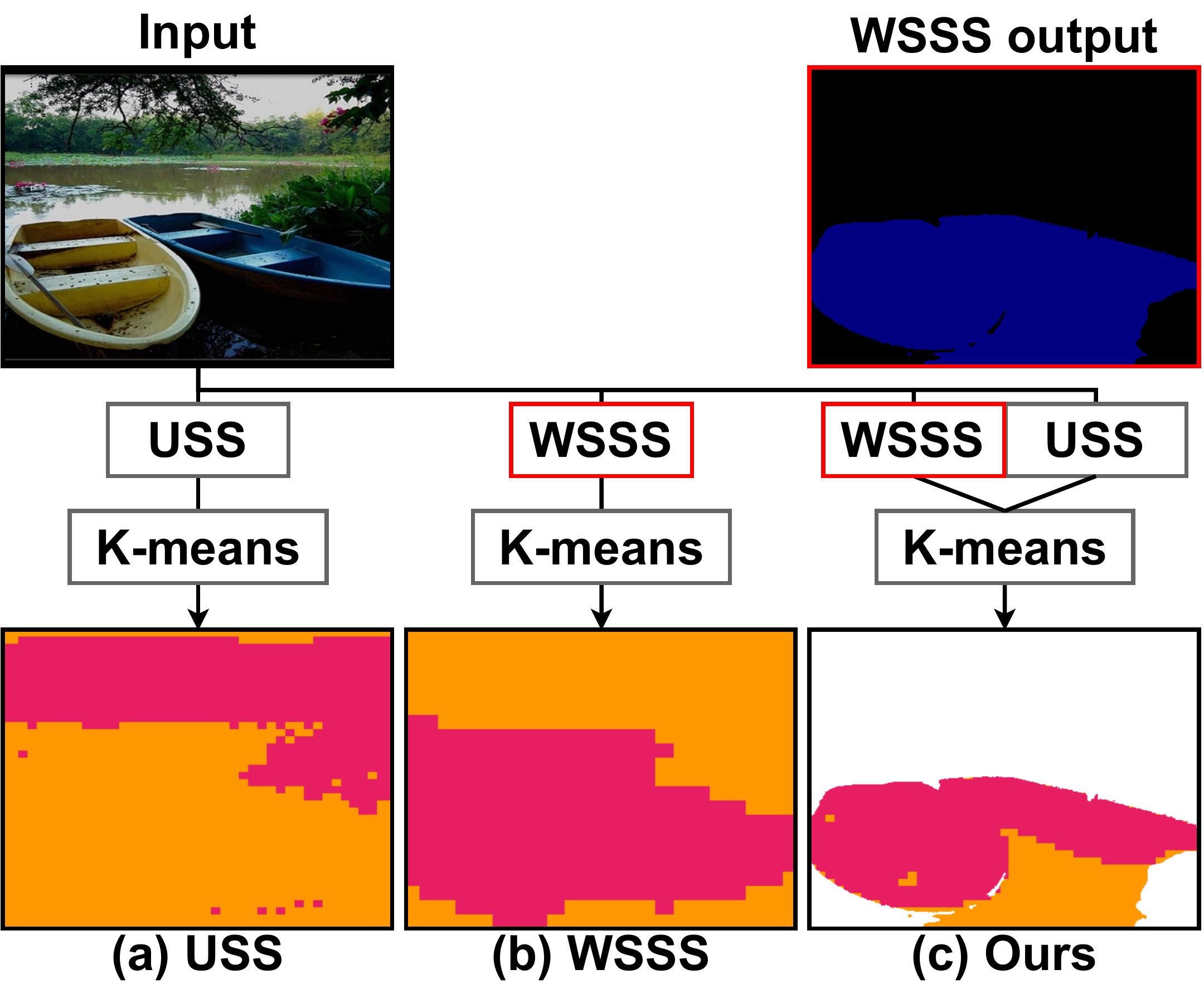}
  \caption{
      Illustration of applying USS into WSSS. \textbf{(a)} and \textbf{(b)}: The simple clustering without the USS or WSSS method cannot separate biased and target objects. \textbf{(c)}: The USS-based clustering separates biased and target objects on a limited area of the WSSS output.
  }
  \label{fig:clustering}
  \vspace{-0.4cm}
\end{figure}

Although two studies \cite{xie2022clims, lee2022weakly} alleviate the biased problem, their requirements hinder WSSS applications in the real world having complex relationships between classes. For example, to apply them to train the Open Images dataset \cite{kuznetsova2020open}, which includes most real-world categories (19,794 classes) in Table \ref{tab:dataset}, they need to not only analyze pairs of the WSSS prediction and image to find biased objects in 6,927 classes (35\% of 19,794 classes) as referred to Fig. \ref{fig:problem}(d) but also confirm the correlation of biased objects and non-problematic classes to prevent decreasing performance of non-problematic classes, impeding the practical WSSS usage. Therefore, without reporting performance on MS COCO 2014 dataset, current debiasing methods \cite{xie2022clims, lee2022weakly} have only shared results on the PASCAL VOC 2012 dataset.

To address the biased problem without additional dataset and supervision, we propose a novel fully-automatic biased removal called MARS (\textbf{M}odel-\textbf{A}gnostic biased object \textbf{R}emoval without additional \textbf{S}upervision), which first utilizes unsupervised semantic segmentation (USS) in WSSS. In particular, our method follows a model-agnostic manner by newly connecting existing WSSS and USS methods for biased removal, which have been only independently studied \cite{jo2022recurseed, hamilton2022unsupervised}. Specifically, our method is based on two key observations related to the integration with USS and WSSS:

\begin{figure}[t]
  \centering
  \includegraphics[width=0.8\linewidth]{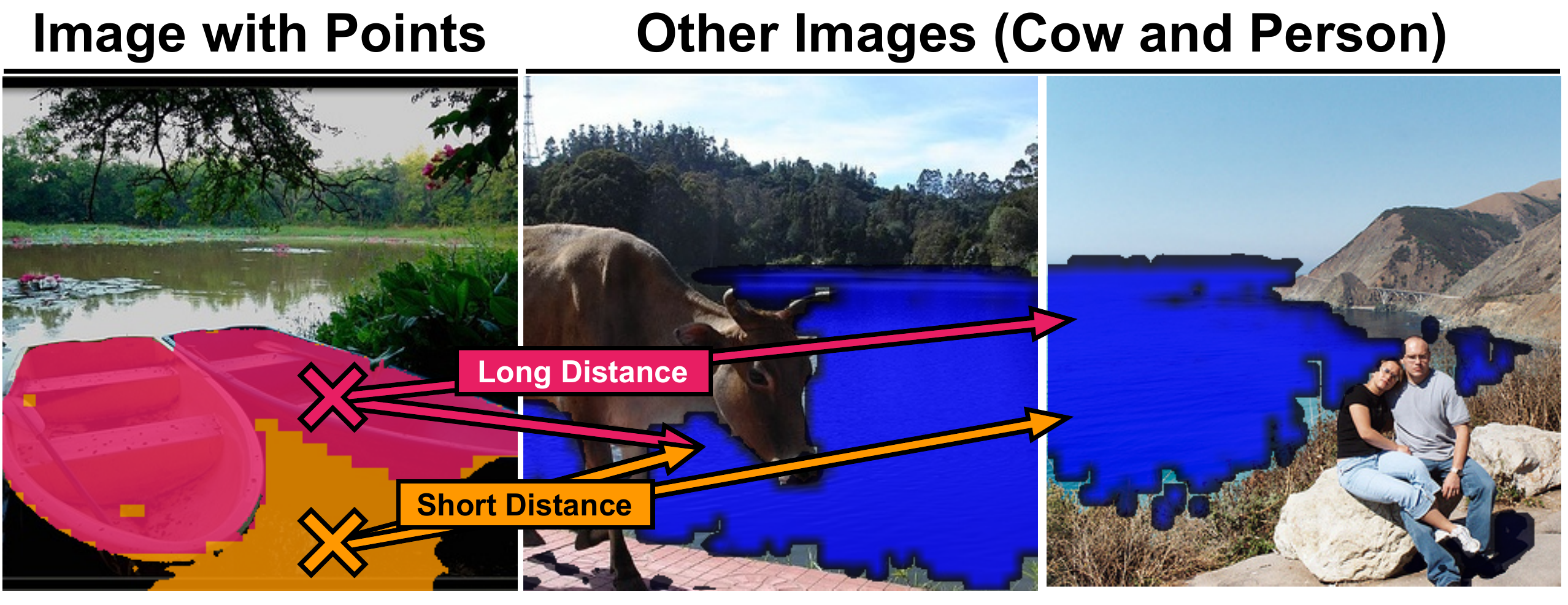}
  \caption{
      Correspondence between biased objects and backgrounds. We measure the distance between each separated object (crosses in the left image) and the background regions of other images (middle and right) within the same dataset. As a result, the long and short distances reflect target and biased objects, respectively. Therefore, the distance of USS features can be used as a criterion to remove biased objects after clustering features. 
  }
  \label{fig:bg_correlation}
  \vspace{-0.4cm}
\end{figure}


\begin{itemize}
\item (The first USS application to separate biased and target objects in WSSS) The USS-based clustering on predicted foreground pixels by the WSSS method successfully disentangles target (pink) and biased (orange) objects, as shown in Fig. \ref{fig:clustering}(c). In contrast, each feature clustering of the WSSS or USS method fails to separate them as illustrated in Figs. \ref{fig:clustering}(a) and (b).
\vspace{-0.2cm}
\item (The first USS-based distance metric to single out the biased object) As shown in Fig. \ref{fig:bg_correlation}, the shorter distance reflects the biased object among distances between two separated regions (pink and orange) and background regions of other images distinguished by the USS method (blue) because the minimum distance between the target and all background sample sets is greater than the minimum distance between the bias and all background sample sets. Accordingly, we show the biased object can exist in the background set, which is a set of classes excluding foreground classes.
\end{itemize}



Therefore, MARS produces debiased labels using the USS-based distance metric after separating biased and target objects in all training images. To prevent increasing FN of non-problematic classes, MARS then complements debiased labels with online predictions in the training time. Our main contributions are summarized as follows. 


\begin{figure*}
  \centering
  \includegraphics[width=1.0\linewidth]{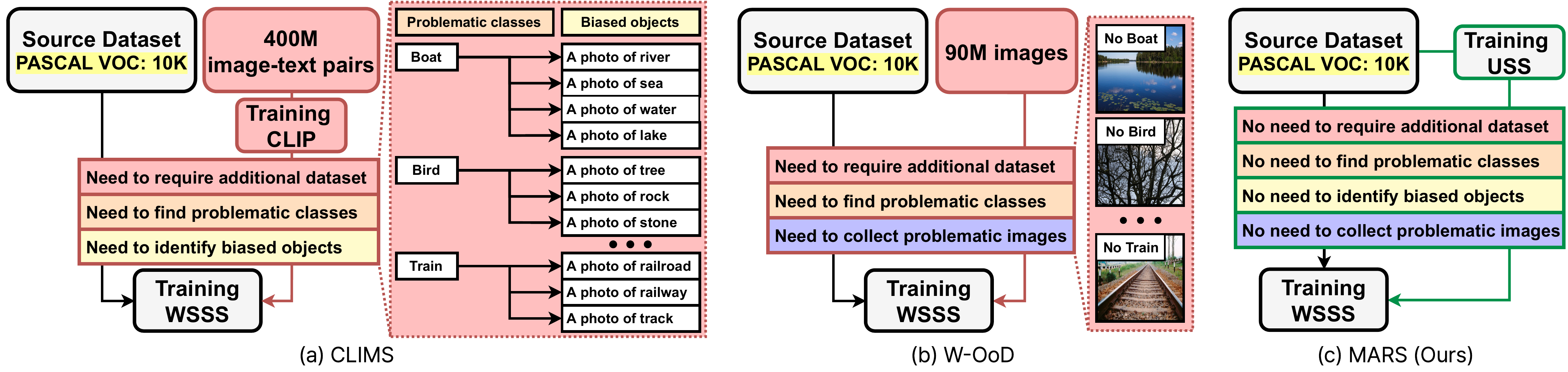}
  \caption{
      Conceptual comparison of three WSSS requirements. \textbf{(a)}: Using the CLIP's knowledge trained on image-text pairs dataset alleviates the biased problem by finding problematic classes and identifying biased objects. \textbf{(b)}: Human annotators manually collect problematic images from the Open Images dataset \cite{kuznetsova2020open} to train biased objects directly. \textbf{(c)}: The proposed MARS first applies an existing USS approach to remove biased objects without additional supervision, achieving the fully-automatic biased removal.
  }
  \label{fig:concept}
  \vspace{-0.4cm}
\end{figure*}

\begin{itemize}
     \item We first introduce two observations of applying USS in WSSS to find biased objects automatically: the USS-based feature clustering for separating biased and target objects and a new distance metric to select the biased object among two isolated objects.
     \item We propose a novel fully-automatic/model-agnostic method, MARS, which leverages semantically consistent features learned through USS to eliminate biased objects without additional supervision and dataset.
     %
     \item Unlike current debiasing methods \cite{xie2022clims, lee2022weakly} that validated only in the PASCAL VOC 2012 dataset with fewer labels, we have also verified the validity of MARS in the more practical case with larger and complex labels such as MS COCO 2014; MARS achieves new state-of-the-art results on two benchmarks (VOC: 77.7\%, COCO: 49.4\%) and consistently improves representative WSSS methods \cite{ahn2019weakly, wang2020self, lee2021anti, jo2022recurseed} by at least 3.4\%, newly validating USS grafting on WSSS.
\end{itemize}

\section{Related Work}\label{sec:relat}

\subsection{Weakly-Supervised Semantic Segmentation}\label{sec:wsss}

Most WSSS approaches \cite{zhang2020reliability, lee2021anti, li2021pseudo, kweon2021unlocking, su2021context, lee2021reducing, qin2022activation, xu2022multi, lee2022threshold} aim to enlarge insufficient foregrounds of initial CAMs. Some studies apply the feature correlation, such as SEAM \cite{wang2020self}, CPN \cite{zhang2021complementary}, PPC \cite{zhang2021complementary}, SIPE \cite{chen2022self}, and RS+EPM \cite{jo2022recurseed}, or patch-based dropout principles, such as FickleNet \cite{lee2019ficklenet}, Puzzle-CAM \cite{jo2021puzzle}, and L2G \cite{jiang2022l2g}. Other methods exploit cross-image information, such as MCIS \cite{sun2020mining}, EDAM \cite{wu2021embedded}, RCA \cite{zhou2022regional}, and $C^2$AM \cite{xie2022c2am}, or global information, such as MCTformer \cite{xu2022multi} and AFA \cite{ru2022learning}. SANCE \cite{li2022towards} and ADELE \cite{liu2022adaptive} propose advanced pipelines to only remove minor noise in pseudo labels. In addition, some studies \cite{lee2021railroad, kim2021discriminative, du2022weakly} employ saliency supervision to remove FP in pseudo labels. However, saliency supervision requires class-agnostic pixel-wise annotations and ignores small and low-prominent objects. All studies mentioned above are independent of our method. We demonstrate consistent improvements of some WSSS approaches \cite{ahn2019weakly, wang2020self, lee2021anti, jo2022recurseed} in Table \ref{tab:wsss}.

\begin{table}
  \caption{  
    Comparison with our method and its related works. With the CLIP model trained on a 400M image-and-text dataset, CLIMS \cite{xie2022clims} removes biased objects after finding problematic classes and identifying biased objects for each class (\emph{i.e.}, a railroad for the train class). W-OoD \cite{lee2022weakly} requires human annotators manually collect problematic images (\emph{i.e.}, only including railroad in an image). Unlike previous approaches, our method removes biased objects without additional datasets and human supervision.
  } 
  \centering
  \begin{scriptsize} 
  \vspace{+0.1cm}
  \begin{tabular}{p{0.21\textwidth} | p{0.07\textwidth} p{0.075\textwidth} p{0.025\textwidth}}
    \toprule
    Properties & CLIMS \protect\cite{xie2022clims} & W-OoD \protect\cite{lee2022weakly} & \textbf{Ours} \\  
    \hline 
    For removing biased objects & \checkmark & \checkmark & \checkmark \\
    Use model-agnostic manner & \xmark & \checkmark & \checkmark \\
    \hline
    \cellcolor{red!25} Need to require additional dataset & \checkmark & \checkmark & \xmark \\
    \cellcolor{orange!25} Need to find problematic classes & \checkmark & \checkmark & \xmark \\
    \cellcolor{yellow!25} Need to identify biased objects & \checkmark & \xmark & \xmark \\
    \cellcolor{blue!25} Need to collect problematic images & \xmark & \checkmark & \xmark \\
    \bottomrule
  \end{tabular}
  \label{tab:novelty}
  \end{scriptsize}
  \vspace{-0.4cm}
\end{table}

\begin{figure*}
  \centering
  \includegraphics[width=1.0\linewidth]{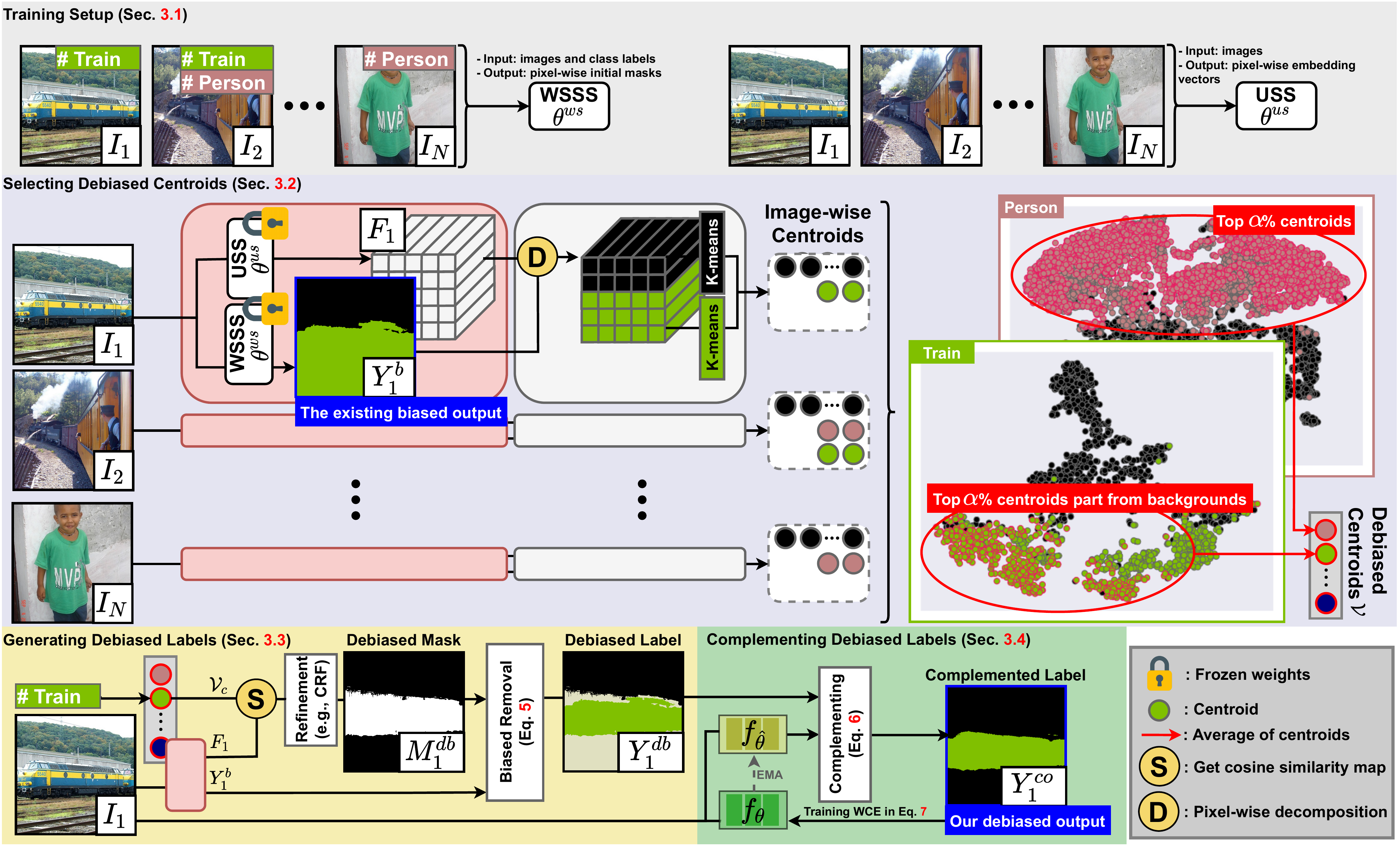}
  \caption{
      Overview of MARS. The USS and WSSS methods, which are trained from scratch, produce pixel-wise embedding vectors $F_{i}$ and the pseudo label $Y^{b}_{i}$, including biased objects, respectively. Based on our observations, K-means clustering generates image-wise centroids (\emph{i.e.}, biased and target objects) from decomposed vectors per class. Then, the debiased centroid $\mathcal{V}^c$ per class is derived as the average of the top $\alpha\%$ centroids from $\{v^c_{i}\}_{i=1}^{N_{c} \cdot K_{fg}}$, the most apart from background centroids of all training images in \eqref{debiased}. To generate the debiased label $Y^{db}_{i}$, we calculate the similarity map using debiased centroids and embedding vectors of the USS method in \eqref{aggregate}. The segmentation network then trains the debiased labels $Y^{db}_{i}$ with the proposed weighted cross-entropy loss function (WCE) in \eqref{wce}. Thus, our MARS provides the final debiased label as $Y^{co}_{i}$.
  }
  \label{fig:overview}
  \vspace{-0.4cm}
\end{figure*}


Similar to our approach, several studies \cite{lee2022weakly, xie2022clims} have focused on removing biased objects in pseudo labels. Table \ref{tab:novelty} compares the essential properties of our method with those of related studies. We also illustrate the conceptual difference with existing WSSS methods \cite{xie2022clims, lee2022weakly} and the proposed MARS in Fig. \ref{fig:concept}. CLIMS \cite{xie2022clims} utilizes the Contrastive Language-Image Pre-training (CLIP) model \cite{radford2021learning}, which is trained on a large-scale dataset of 400 million image-text pairs (\emph{i.e.}, using text supervision), and needs to identify biased objects (\emph{e.g.}, railroad and sea) in all problematic classes (\emph{e.g.}, train and boat classes), as shown in Fig. \ref{fig:problem}(d). W-OoD \cite{lee2022weakly} needs human annotators to collect additional images, which only include biased objects (\emph{e.g.}, railroad and sea), from the Open Images dataset \cite{kuznetsova2020open} to train the classification network directly with problematic images. Our method first removes biased objects by leveraging the semantic consistency of the trained USS method from scratch without additional human supervision and dataset. 

\subsection{Unsupervised Semantic Segmentation}\label{sec:uss}

USS focuses on training semantically meaningful features within image collection without any form of annotations. Therefore, all USS methods \cite{caron2018deep, ji2019invariant, ouali2020autoregressive, cho2021picie, van2021unsupervised, van2022discovering, ziegler2022self, hamilton2022unsupervised} are used as the pre-training strategy because they cannot produce class-aware predictions only by grouping features. IIC \cite{ji2019invariant}, AC \cite{ouali2020autoregressive}, and PiCIE \cite{cho2021picie} maximize the mutual information between different views. Leopart \cite{ziegler2022self}, and STEGO \cite{hamilton2022unsupervised} utilize the self-supervised vision transformer to learn spatially structured image representations, resulting in accurate object masks without additional supervision. Notably, STEGO \cite{hamilton2022unsupervised} enriches correlations between unsupervised features with training a simple feed-forward network, leading to efficient training without re-training or fine-tuning weights initialized by DINO \cite{caron2021emerging}. Our method is agnostic to the underlying USS methods, utilizing pixel-wise semantic features only. Hence, all USS methods are independent of our approach. We show consistent improvements in recent USS methods \cite{ziegler2022self, hamilton2022unsupervised}, verifying the flexibility of our method and the potential for integrating future advances in USS into our method.


\section{Method}\label{sec:metho}



The proposed MARS consists of four sections/stages: (a) training WSSS and USS methods for the model-agnostic manner, (b) selecting debiased centroids, (c) generating debiased labels, and (d) complementing debiased labels during the learning process. 
The overall framework of MARS is illustrated in Fig. \ref{fig:overview}.

\subsection{Training Setup} \label{sec:setup}

This section describes the training setup for existing WSSS and USS models. Unlike \cite{xie2022clims, lee2022weakly}, our model-agnostic approach does not require additional datasets for training these models. For a fair comparison, we train all WSSS and USS models from scratch on the PASCAL VOC 2012 or MS COCO 2014 datasets, following the standard setup of WSSS methods \cite{ahn2019weakly, wang2020self, lee2021anti, jo2022recurseed}. Each training image $I_{i} \in \mathbb{R}^{3 \times H \times W}$ in the dataset is associated with a set of image-level class labels $L_{i} \in \{0, 1\}^{C}$, where $C$ is the number of categories/classes. In detail, the classification network generates initial CAMs after training images and image-level class labels. Then, the conventional propagating method \cite{ahn2019weakly} refines initial CAMs to produce pseudo labels. Finally, USS methods \cite{ziegler2022self, hamilton2022unsupervised} are trained only on the images, following each pretext task. For the following sections, our method utilizes pseudo masks and semantic features produced from the frozen weights of the WSSS and USS methods, respectively.

\subsection{Selecting Debiased Centroids} \label{sec:selecting}

This section describes how our approach separates biased and target objects using trained WSSS and USS methods. For a mini-batch image $I_{i}$, the trained USS method generates pixel-wise embedding vectors $F_{i} \in \mathbb{R}^{D \times H \times W}$, not including class-specific information. Meanwhile, the trained WSSS method produces pseudo labels $Y^{b}_{i} \in \{0,1,...,C\}^{H \times W}$, including both biased and target objects. We group pixel-wise embedding vectors $F_{i}$ under $Y^{b}_{i}$'s prediction region $\{(y,x)|Y^{b}_{i}(y,x)=c\}$ for each class $c$, and apply K-means clustering to generate image-wise centroids $v^{c}_{i \cdot K + j} \in \mathbb{R}^{D}$ per class $c$ for $j\in\{1,2,...,K\}$. Here, the number $K$ of clusters for foreground ($c>0$) and background ($c=0$) classes are $K_{fg}$ and $K_{bg}$, respectively. We set $K_{fg}$ to 2 to separate biased and target objects, and $K_{bg}$ can be varied. 
Although our aforementioned simple clustering isolates biased and target objects, it cannot identify which one is the target or biased object among both candidate objects. To single out the biased object, we propose a new following distance metric between each candidate object and background centroids in all training images in \eqref{distance}:
\begin{align} \label{distance}
dist^{c}_{k} = \frac{1}{N^{bg}} \sum_{j=0}^{N^{bg}} D(v^{c}_{k}, v^{0}_{j}) 
\end{align} 
where $0$ and $c$ denote the index of the background and foreground classes, respectively, $i$ denotes the index of the foreground centroid, and $N^{bg}:= N \cdot K_{bg}$ denotes the number of background centroids from all $N$ training images. $S(\cdot)$ and $D(\cdot)$ mean the cosine similarity (i.e., ${v \cdot v'} / {\left\| v \right\|\left\| v' \right\|}$) and distance (i.e., $(1- S(v, v'))/2$), respectively. For instance, long and short distances mean target and biased centroids, respectively, since each distance reflects the degree of whether to include the biased object as shown in Fig. \ref{fig:bg_correlation}. We sort all foreground centroids per class in descending order by the distance using background centroids. Thus, for each class $c$, we aggregate the average of top $\alpha\%$ centroids most apart from background centroids to get a single vector representing the final debiased/target centroid $\mathcal{V}^{c} \in \mathbb{R}^{D}$ as follows: 
\begin{align} \label{debiased}
& \mathcal{V}^{c} = \frac{1}{{\left \lceil N^{fg}_{c} \cdot \alpha \right \rceil}}\sum_{j\in\{k_1,k_2,...,k_{\left \lceil N^{fg}_{c} \cdot \alpha \right \rceil}\}} v^{c}_{j}, \\ 
& dist^{c}_{k_1} \geq dist^{c}_{k_2} \geq ... \geq dist^{c}_{k_{N^{fg}_{c}}} \label{index}
\end{align}
where $N^{fg}_{c}:= N_{c} \cdot K_{fg}$ denotes the number of centroids from $N_{c}$ images having class $c$, $\alpha \in [0, 1]$ is the ratio of selecting target centroids, and $\{k_i\}_{i\in\{1:N^{fg}_{c}\}}$ is the ordered index set satisfying \eqref{index} (e.g., $v^{c}_{k_{1}}$ is the centroid having the largest distance from all background centroids). 
In other words, when we identify the biased or target/debiased object in the given image $I_i$, we improve its identification performance by using information from other training images together; its analysis is detailed in Sec. \ref{sec:analysis}.

\subsection{Generating Debiased Labels} \label{sec:generating}

We present our approach for finding and removing biased pixels in pseudo labels $Y^{b}_{i}$. We first compute the similarity map between each debiased centroid $\mathcal{V}^{c}$ and embedding vectors $F_{i}$ for per-pixel biased removal. However, we observe that the trained USS method cannot separate some classes if two categories (e.g., horse and sheep) have the same super-category (e.g., animals). This issue is also present in current USS methods \cite{cho2021picie, ziegler2022self, hamilton2022unsupervised} and is caused by the inability to distinguish between objects within the same super-category. To address this shortcoming, we introduce a debiasing process that generates the debiased mask $\hat{M}^{db}_{i}$ using the pixel-wise maximum function as follows: 
\begin{align} \label{aggregate}
&  \hat{M}^{db}_{i}(y, x) = ReLU\Big( \max_{c \in \mathcal{C}_{I_i}} S(F_{i}[:, y, x], \mathcal{V}^{c}) \Big)
\end{align}
where ($x,y$) indicates $x,y$-th pixel position, $F_{i}(:, y, x) \in \mathbb{R}^{D}$ is the pixel-wise embedding vector, $\mathcal{V}^{c} \in \mathbb{R}^{D}$ denotes the debiased/target centroid for each class $c$, $\mathcal{C}_{I_i}$ is corresponding class indices of each image $I_i$, and the ReLU activation removes negative values in $\hat{M}^{db}_{i} \in [-1, 1]^{H \times W}$. After applying a typical post-processing refinement (\emph{e.g.}, CRF \cite{krahenbuhl2011efficient}) to $\hat{M}^{db}_{i}$, we generate the binary debiased mask $M^{db}_{i} \in \{0, 1\}^{H \times W}$, which produces the debiased label $Y^{db}_{i} = \{-1,0,1,...,c\}^{H \times W}$ using the binary debiased mask $M^{db}_{i}$ and the WSSS label $Y^{b}_{i}$ as follows:
\begin{small}
\begin{equation} \label{removal}
    Y^{db}_{i}(y,x) = 
        \begin{cases*}
            -1              , & if $Y^{b}_{i}(y,x) > 0$ and $M^{db}_{i}(y,x) = 0$, \\
            Y^{b}_{i}(y,x)  , & \text{otherwise} \\
        \end{cases*}
\end{equation}
\end{small}
where $-1$ indicates the new biased class for the next section \ref{sec:complementing}. The pixel value in the debiased label $Y^{db}_{i}$ is only replaced with the biased class ($-1$) if our debiased mask $M^{db}_{i}$ and the WSSS mask $Y^{b}_{i}$ provide the label 0 and the foreground class ($> 0$), respectively. Namely, we remove biased predictions of WSSS by computing the per-pixel similarity of debiased centroids within the embedding space.

\subsection{Complementing Debiased Labels} \label{sec:complementing} 
\begin{figure}[t]
  \centering
\includegraphics[width=0.9\linewidth]{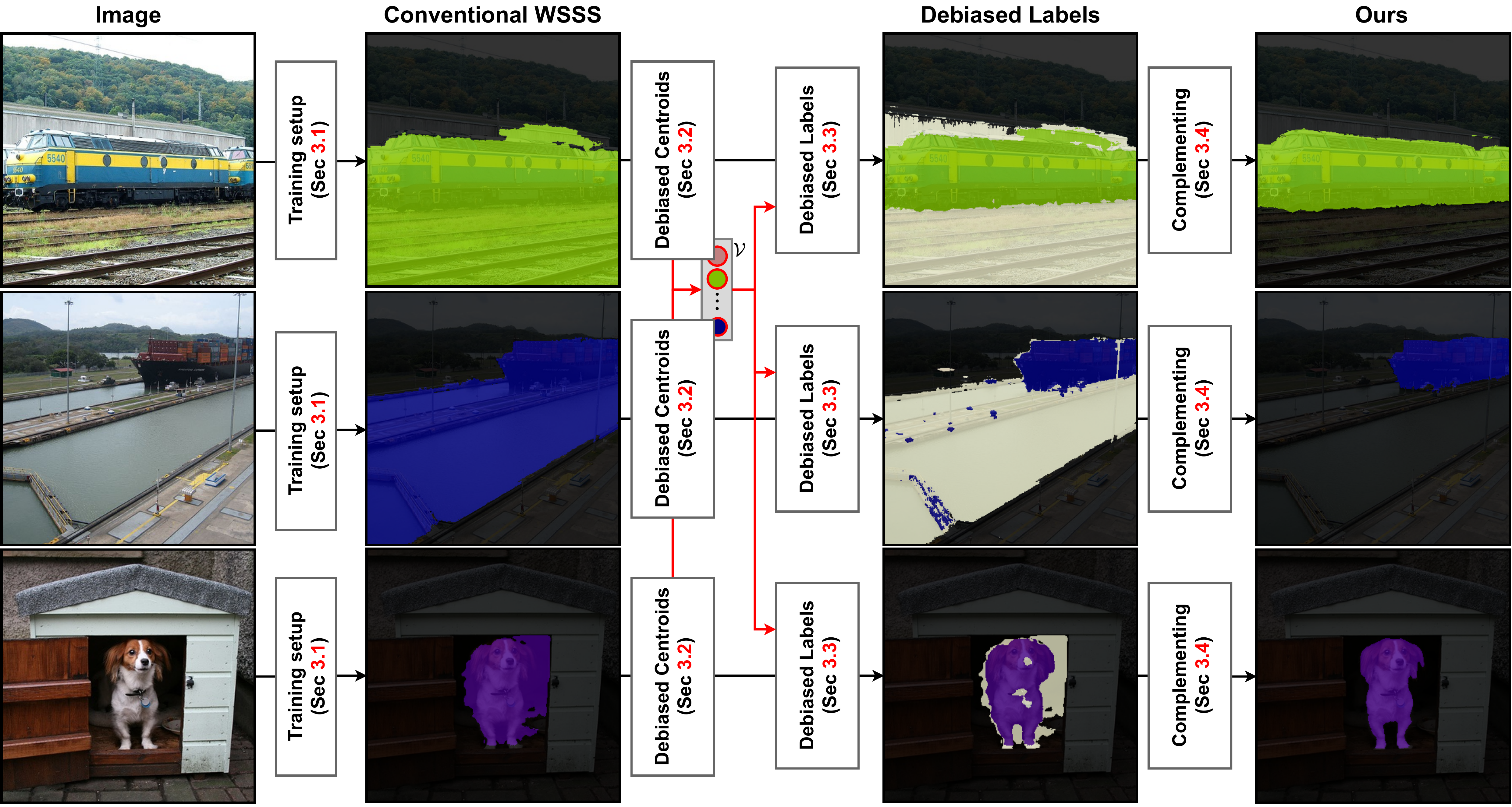}
  \caption{
      Effect of the proposed components. 
      For problem classes including the biased objects, \emph{e.g.}, boat and train classes, second and third components (Secs. \ref{sec:selecting} and \ref{sec:generating}) effectively remove biased objects in debiased labels $Y^{db}_{i}$ and then the fourth component (Sec. \ref{sec:complementing}) preserves removed objects (the first and second samples). For non-problematic classes not containing biased objects, \emph{e.g.}, the dog class, the fourth component accurately restores non-biased objects (the third sample). In addition, the red line denotes applying debiased centroids to produce debiased labels. 
  }
  \label{fig:effect}
  \vspace{-0.4cm}
\end{figure}

This last section proposes a new training strategy to complement biased pixels in debiased labels. As shown in Fig. \ref{fig:effect}, although biased objects in our debiased labels are successfully removed for problematic classes (\emph{i.e.}, classes including biased objects, \emph{e.g.}, train and boat classes (the first and second images)), we observe non-biased objects (\emph{e.g.}, people's clothes, eyes of animals, wheels of vehicles) are also eliminated, increasing FN of non-problematic classes, \emph{e.g.}, the dog class (the third image). To complement non-biased objects, we utilize online predictions $\hat{P_{i}}$ from a teacher network during its learning process with certain masks.

\begin{figure}[t]
  \centering
  \includegraphics[width=0.9\linewidth]{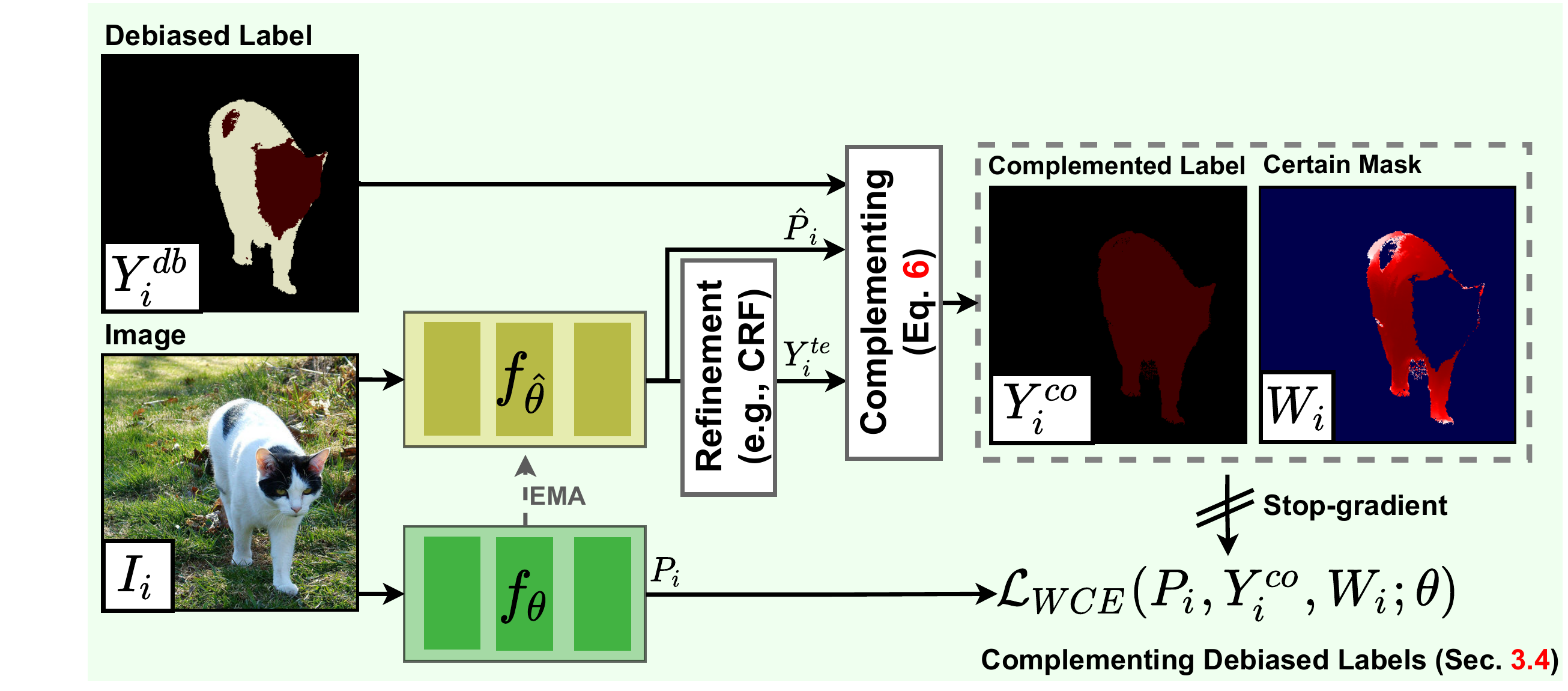}
  \caption{
      Illustration of the proposed complementing process. With the refinement, the teacher network produces the teacher's label $Y^{te}_{i}$. To prevent increasing FN of non-problematic classes, biased pixels in debiased labels $Y^{db}_{i}$ are complemented with the teacher's prediction. To avoid training uncertain labels, the student network is updated using the proposed WCE in \eqref{wce} with complemented labels $Y^{co}_{i}$ and certain masks $W_{i}$, resulting in the final predictions similar to ground truths.
  }
  \label{fig:complementing}
  \vspace{-0.4cm}
\end{figure}
We illustrate the complementing process as shown in Fig. \ref{fig:complementing}. Here, $\theta$ denotes weights of the student network, and we update a teacher network $\hat{\theta}$ using an exponential moving average (EMA). The student and teacher networks predict segmentation outputs $P_{i},\hat{P}_{i} \in [0, 1]^{C \times H \times W}$ after applying the softmax function. We then employ the refinement $R$ (\emph{e.g.}, CRF \cite{krahenbuhl2011efficient}) and $\argmaxB$ operator to produce the teacher's label $Y^{te}_{i} = \{0,1,...,c\}^{H \times W}$. Finally, we generate complemented labels $Y^{co}_{i} \in \{0,1,...,c\}^{H \times W}$ by filling biased classes ($-1$) in debiased labels $Y^{db}_{i} \in \{-1,0,1,...,c\}^{H \times W}$ with the teacher's prediction $Y^{te}_{i}$.

However, when updating the teacher network in early epochs, the complemented label $Y^{co}_{i}$ includes incorrect predictions in smooth probabilities (\emph{i.e.}, uncertain predictions), covering biased objects in the complementing process. To address this issue in uncertain pixels, we propose a concept of a certain mask $W_{i} \in [0, 1]^{H \times W}$, which is the matrix of pixel-wise maximum probabilities for all foreground classes, and its ablation analysis is detailed in Sec. \ref{sec:analysis}:
\begin{small}
\begin{equation} \label{certainty}
    W_{i}(y,x) = 
        \begin{cases*}
            \underset{c \in \mathcal{C}_{I_i}}{\max} \hat{P}_{i}({c, y, x}), & if $Y^{db}_{i}(y, x) = -1$, \\
            1, & \text{otherwise}
        \end{cases*}
\end{equation}
\end{small}
where $\mathcal{C}_{I_i}:=\{k\,|\,L_{i}(k)=1\}$ is an index set of truth classes for each image $I_i$ and $-1$ denotes the complemented/biased class. To train the segmentation network with complemented labels $Y^{co}_{i}$ and certain masks $W_{i}$, we propose the weighted cross entropy (WCE) loss that multiplies the certain mask $W_{i}$ with the per-pixel cross-entropy loss to reflect the uncertainty ratio:
\begin{small}
\begin{align} \label{wce}
& \mathcal{L}_{WCE}(P_{i},Y^{co}_{i},W_{i}; \theta)\\\nonumber 
&= - \sum_{c \in \mathcal{C}} \sum_{{y, x} \in \mathcal{W}} W_{i}(y, x) \cdot O[Y^{co}_{i}](c, y, x) \log P^{\theta}_{i}(c, y, x) 
\end{align}
\end{small}
where $O[\cdot]$ means one-hot encoding for the per-pixel cross-entropy loss function. As a result, the proposed MARS successfully removes biased objects without performance degradation of non-problematic classes by complementing biased pixels in debiased labels with the teacher's predictions in its learning process (the bottom results in Fig. \ref{fig:effect}).

In summary, Fig. \ref{fig:effect} illustrates the effect of the proposed components on the WSSS performance, following examples in Fig. \ref{fig:problem}(c) (see examples of other classes in Appendix): After training WSSS and USS methods in Sec. \ref{sec:setup}, the first component (Sec. \ref{sec:selecting}) extracts debiased centroids $\{\mathcal{V}^{c}\}_{c=1}^C$  based on the distance of all background centroids to each foreground centroid. The second component (Sec. \ref{sec:generating}) generates debiased labels $Y^{db}_{i}$ using debiased centroids and previous WSSS labels. The last component (Sec. \ref{sec:complementing}) trains the segmentation network by complementing biased pixels to provide the final debiased label as $Y^{co}_{i}$. We provide a detailed analysis of our method in Sec. \ref{sec:analysis}.

\section{Experiments}\label{sec:exper}

\subsection{Experimental Setup}


\textbf{Datasets.} We conduct all experiments on the PASCAL VOC 2012 \cite{everingham2010pascal} and MS COCO 2014 \cite{lin2014microsoft} datasets, both of which contain image-level class labels, bounding boxes, and pixel-wise annotations. Despite the difficulty of MS COCO 2014 dataset \cite{lin2014microsoft}, \emph{e.g.}, small-scale objects and imbalance class labels, our method significantly improves all benchmarks. PASCAL VOC 2012 \cite{everingham2010pascal} and MS COCO 2014 \cite{lin2014microsoft} datasets have 21 and 81 classes, respectively. 



\textbf{Implementation details.} To ensure a fair comparison with existing methods, we train two USS methods \cite{liu2022adaptive, lee2022weakly} from scratch on each dataset. To demonstrate the scalability of our method, we utilize four WSSS methods \cite{ahn2019weakly, wang2020self, lee2021anti, jo2022recurseed} on PASCAL VOC 2012 dataset \cite{everingham2010pascal}. All WSSS and USS methods' hyperparameters and architectures are the same as those in their respective papers. Thus, our method has the same runtime as other methods in evaluation. We only use two hyperparameters to select debiased centroids: $K_{bg}$ is set to 2, and $\alpha$ is set to 0.40. In addition, we use multi-scale inference and CRF \cite{krahenbuhl2011efficient} with conventional settings to evaluate the segmentation network's performance. We conduct all experiments on a single RTX A6000 GPU and implement all WSSS/USS methods in PyTorch.


\textbf{Evaluation metrics.} We evaluate our method using mIoU, following the typical evaluation metric of existing WSSS studies \cite{ahn2018learning, ahn2019weakly, wang2020self, lee2021anti, jo2022recurseed}. We also follow FP and FN metrics proposed by the gold standard \cite{wang2020self}. We obtain all results for the PASCAL VOC 2012 \emph{val} and \emph{test} sets from the official PASCAL VOC online evaluation server.

\begin{table}[t]
    \centering
    \caption{ 
    Performance comparison of WSSS methods regarding mIoU ($\%$) on PASCAL VOC 2012 and COCO 2014. {* and $\dagger$ indicate the backbone of VGG-16 and ResNet-50, respectively.} Sup., supervision; $\mathcal{I}$, image-level class labels; $\mathcal{S}$, saliency supervision; $\mathcal{D}$, using the external dataset; $\mathcal{F}$, pixel-wise annotations (i.e, fully-supervised semantic segmentation). 
  }
  \vspace{+0.1cm}
  \begin{scriptsize}
  \begin{tabular}{p{0.150\textwidth} c c c c c}
    \toprule
    \multirow{2}{*}{\begin{tabular}[c]{@{}c@{}}Method\end{tabular}} & 
    \multirow{2}{*}{\begin{tabular}[c]{@{}c@{}}Backbone\end{tabular}} & 
    \multirow{2}{*}{\begin{tabular}[c]{@{}c@{}}Sup.\end{tabular}} & \multicolumn{2}{c}{VOC}  & COCO \\
           &          &      & \emph{val} & \emph{test} & \emph{val} \\
    \hline \hline
    DSRG {\tiny CVPR'18} \cite{huang2018weakly} & R101 & $\mathcal{I}$+$\mathcal{S}$ & 61.4 & 63.2 & 26.0* \\
    FickleNet {\tiny CVPR'19} \cite{lee2019ficklenet} & R101 & $\mathcal{I}$+$\mathcal{S}$ & 64.9 & 65.3 & - \\
    MCIS {\tiny ECCV'20} \cite{sun2020mining} & R101 & $\mathcal{I}$+$\mathcal{S}$ & 66.2 & 66.9 & - \\
    CLIMS {\tiny CVPR'22} \cite{xie2022clims} & R50 & $\mathcal{I}$+$\mathcal{D}$ & 69.3 & 68.7 & - \\
    W-OoD {\tiny CVPR'22} \cite{lee2022weakly} & R101 & $\mathcal{I}$+$\mathcal{D}$ & 69.8 & 69.9 & - \\
    EDAM {\tiny CVPR'21} \cite{wu2021embedded} & R101 & $\mathcal{I}$+$\mathcal{S}$ & 70.9 & 70.6 & - \\
    EPS {\tiny CVPR'21} \cite{lee2021railroad} & R101 & $\mathcal{I}$+$\mathcal{S}$ & 70.9 & 70.8 & 35.7* \\
    DRS {\tiny AAAI'21} \cite{kim2021discriminative} & R101 & $\mathcal{I}$+$\mathcal{S}$ & 71.2 & 71.4 & - \\
    L2G {\tiny CVPR'22} \cite{jiang2022l2g} & R101 & $\mathcal{I}$+$\mathcal{S}$ & 72.1 & 71.7 & 44.2 \\
    RCA {\tiny CVPR'22} \cite{zhou2022regional} & R101 & $\mathcal{I}$+$\mathcal{S}$ & 72.2 & 72.8 & 36.8* \\
    PPC {\tiny CVPR'22} \cite{du2022weakly} & R101 & $\mathcal{I}$+$\mathcal{S}$ & 72.6 & 73.6 & - \\
    \hline
    PSA {\tiny CVPR'18} \cite{ahn2018learning} & WR38 & $\mathcal{I}$ & 61.7 & 63.7 & - \\
    IRNet {\tiny CVPR'19} \cite{ahn2019weakly} & R50 & $\mathcal{I}$ & 63.5 & 64.8 & - \\
    SSSS {\tiny CVPR'20} \cite{araslanov2020single} & WR38 & $\mathcal{I}$ & 62.7 & 64.3 & - \\
    RRM {\tiny AAAI'20} \cite{zhang2020reliability} & R101 & $\mathcal{I}$ & 66.3 & 65.5 & - \\
    SEAM {\tiny CVPR'20} \cite{wang2020self} & WR38 & $\mathcal{I}$ & 64.5 & 65.7 & 31.9 \\
    CDA {\tiny ICCV'21} \cite{su2021context} & WR38 & $\mathcal{I}$ & 66.1 & 66.8 & 33.2 \\
    AdvCAM {\tiny CVPR'21} \cite{lee2021anti} & R101 & $\mathcal{I}$ & 68.1 & 68.0 & - \\
    CSE {\tiny ICCV'21} \cite{kweon2021unlocking} & WR38 & $\mathcal{I}$ & 68.4 & 68.2 & 36.4 \\
    ReCAM {\tiny CVPR'22} \cite{chen2022class} & R101 & $\mathcal{I}$ & 68.5 & 68.4 & - \\
    CPN {\tiny ICCV'21} \cite{zhang2021complementary} & WR38 & $\mathcal{I}$ & 67.8 & 68.5 & - \\
    RIB {\tiny NeurIPS'21} \cite{lee2021reducing} & R101 & $\mathcal{I}$ & 68.3 & 68.6 & 43.8 \\
    ADELE {\tiny CVPR'22} \cite{liu2022adaptive} & WR38 & $\mathcal{I}$ & 69.3 & 68.8 & - \\
    PMM {\tiny ICCV'21} \cite{li2021pseudo} & WR38 & $\mathcal{I}$ & 68.5 & 69.0 & 36.7 \\
    AMR {\tiny AAAI'22} \cite{qin2022activation} & R101 & $\mathcal{I}$ & 68.8 & 69.1 & - \\
    URN {\tiny AAAI'22} \cite{li2022uncertainty} & R101 & $\mathcal{I}$ & 69.5 & 69.7 & 40.7 \\ 
    SIPE {\tiny CVPR'22} \cite{chen2022self} & R101 & $\mathcal{I}$ & 68.8 & 69.7 & 40.6 \\
    AMN {\tiny CVPR'22} \cite{lee2022threshold} & R101 & $\mathcal{I}$ & 69.5 & 69.6 & 44.7 \\
    MCTformer {\tiny CVPR'22} \cite{xu2022multi} & WR38 & $\mathcal{I}$ & 71.9 & 71.6 & 42.0 \\
    SANCE {\tiny CVPR'22} \cite{li2022towards} & R101 & $\mathcal{I}$ & 70.9 & 72.2 & 44.7$\dagger$ \\
    RS+EPM {\tiny Arxiv'22} \cite{jo2022recurseed} & R101 & $\mathcal{I}$ & 74.4 & 73.6 & 46.4 \\
    \rowcolor{maroon!25} MARS (Ours) & R101 & $\mathcal{I}$ & \textbf{77.7} & \textbf{77.2} & \textbf{49.4} \\
    FSSS & R101 & $\mathcal{F}$ & 80.6 & 81.0 & 61.8 \\ 
    \bottomrule
  \end{tabular} 
  \label{tab:performance}
  \end{scriptsize}
  \vspace{-0.2cm}
\end{table}

\subsection{Comparison with state-of-the-art approaches} 


We compare our method with other WSSS methods in Table \ref{tab:performance}. Recent state-of-the-art methods exploit additional supervision to reduce the number of FP in pseudo labels, such as saliency supervision \cite{hou2017deeply, liu2019simple, pang2020multi}, the external dataset to collect biased images \cite{lee2022weakly}, and text supervision from an image-to-text dataset (\emph{e.g.}, CLIP \cite{radford2021learning}). By contrast, without additional supervision and dataset, we mitigate the biased problem by leveraging the inherent advantage of USS, outperforming previous state-of-the-art methods by at least 3.3\%. We also refer to Appendix for the qualitative comparison with existing WSSS methods and ours.





\subsection{Analysis} \label{sec:analysis}

\textbf{Flexibility.} We demonstrate the flexibility of our method by comparing it to various WSSS and USS methods. As shown in Table \ref{tab:uss}, our method consistently outperforms existing WSSS methods regardless of applying Leopart \cite{ziegler2022self} or STEGO \cite{hamilton2022unsupervised} for our method. In Table \ref{tab:wsss}, we compare our method to two flexible WSSS methods \cite{liu2022adaptive, lee2022weakly} based on four WSSS methods \cite{ahn2019weakly, wang2020self, lee2021anti, jo2022recurseed}. For the WSSS experiment, we utilize STEGO \cite{hamilton2022unsupervised} because this USS method performs best in Table \ref{tab:uss}. We employ the same backbone and segmentation model to ensure a fair comparison. Surprisingly, our method improves each performance by 6.3\%, 6.3\%, 2.2\%, and 3.3\% for IRNet \cite{ahn2019weakly}, SEAM \cite{wang2020self}, AdvCAM \cite{lee2021anti}, and RS+EPM \cite{jo2022recurseed}, respectively, as shown in Table \ref{tab:wsss}. The qualitative improvements with ADELE \cite{liu2022adaptive}, W-OoD \cite{lee2022weakly}, and ours are given in Appendix. Although W-OoD \cite{lee2022weakly} addresses the biased problem, it requires the manual collection of images, only including biased objects from an additional dataset (\emph{e.g.}, Open Images \cite{kuznetsova2020open}). The proposed MARS first removes biased objects without additional human supervision, verifying the flexibility and superiority of our method.

\begin{table}
    \centering
  \caption{ 
    Comparison with two USS methods \cite{ziegler2022self, hamilton2022unsupervised} in terms of mIoU (\%) on PASCAL VOC 2012 dataset.
  }
  \vspace{+0.1cm}
  \begin{scriptsize}
  \begin{tabular}{p{0.100\textwidth} >{\centering}p{0.08\textwidth} >{\centering}p{0.05\textwidth} | c c}
    \toprule
    Method      & USS & Backbone     & mIoU (\emph{val}) & mIoU (\emph{test}) \\
    \hline \hline
    IRNet \cite{ahn2019weakly} & \xmark & R50 & 63.5       & 64.8 \\
    + Ours & Leopart \cite{ziegler2022self} & R50 & 68.1 & 68.8 \\
    \rowcolor{maroon!25} + Ours & STEGO \cite{hamilton2022unsupervised} & R50 & \textbf{69.8} & \textbf{70.9} \\
    \hline
    RS+EPM \cite{jo2022recurseed} & \xmark & R101         & 74.4       & 73.6 \\
    + Ours & Leopart \cite{ziegler2022self} & R101 & 75.4 & 75.8 \\
    \rowcolor{maroon!25} + Ours & STEGO \cite{hamilton2022unsupervised} & R101 & \textbf{77.7} & \textbf{77.2} \\
    \hline
    \bottomrule
  \end{tabular}
  \label{tab:uss}
  \end{scriptsize}
  \vspace{-0.2cm}
\end{table}

\begin{table}
    \centering
  \caption{ 
    Comparison with four WSSS methods \cite{ahn2019weakly, wang2020self, lee2021anti, jo2022recurseed} in terms of mIoU (\%) on PASCAL VOC 2012 dataset. FSSS means training the dataset with pixel-wise annotations. $(\cdot)$ means the percentage improvement in the gap between WSSS and FSSS.
  }
  \vspace{+0.1cm}
  \begin{scriptsize}
  \begin{tabular}{p{0.085\textwidth} >{\centering}p{0.050\textwidth} >{\centering}p{0.09\textwidth} | c c}
    \toprule
    Method      & Backbone & Segmentation     & mIoU (\emph{val}) & mIoU (\emph{test}) \\
    \hline \hline
    IRNet \cite{ahn2019weakly} & R50 & DeepLabv2 & 63.5       & 64.8 \\
   \rowcolor{maroon!25} + Ours & R50 & DeepLabv2 & \textbf{69.8 (49\%)} & \textbf{70.9 (52\%)} \\
   FSSS & R50 & DeepLabv2 & 76.3 & 76.5 \\
   \hline
    SEAM \cite{wang2020self} & WR38 & DeepLabv1         & 64.5       & 65.7 \\
    + ADELE \cite{liu2022adaptive} & WR38 & DeepLabv1         & 69.3 (35\%)      & 68.8 (25\%) \\
   \rowcolor{maroon!25} + Ours & WR38 & DeepLabv1 & \textbf{70.8 (46\%)} & \textbf{71.4 (46\%)} \\
   FSSS & WR38 & DeepLabv1 & 78.1 & 78.2 \\
   \hline
   AdvCAM \cite{lee2021anti} & R101 & DeepLabv2         & 68.1       & 68.0 \\
    + W-OoD \cite{lee2022weakly} & R101 & DeepLabv2         & 69.8 (17\%)      & 69.9 (18\%) \\
   \rowcolor{maroon!25} + Ours & R101 & DeepLabv2 & \textbf{70.3 (22\%)} & \textbf{71.2 (30\%)} \\
   FSSS & R101 & DeepLabv2 & 78.0 & 78.6 \\
   \hline
   RS+EPM \cite{jo2022recurseed} & R101 & DeepLabv3+         & 74.4       & 73.6 \\
   \rowcolor{maroon!25} + Ours & R101 & DeepLabv3+ & \textbf{77.7 (53\%)} & \textbf{77.2 (49\%)} \\
   FSSS & R101 & DeepLabv3+ & 80.6 & 81.0 \\
   \hline
    \bottomrule
  \end{tabular}
  \label{tab:wsss}
  \end{scriptsize}
  \vspace{-0.2cm}
\end{table}

\begin{table}
  \centering
  \caption{ 
    Effect of key components in terms of mIoU (\%) on PASCAL VOC 2012 \emph{train} set.
  }
  \vspace{+0.1cm}
  \begin{scriptsize}
  \begin{tabular}{p{0.0001\textwidth} c c | p{0.05\textwidth} p{0.07\textwidth} p{0.07\textwidth}}
   \toprule
      & Complementing & WCE \eqref{wce} & mIoU & FP & FN \\ 
    \hline \hline
    1 & \xmark & \xmark & 77.4 & 0.123 & 0.108 \\
    2 & \checkmark & \xmark & 80.9 & 0.122 & 0.075 \\
    \rowcolor{maroon!25} 3 & \checkmark & \checkmark & \textbf{81.8} & \textbf{0.099} & \textbf{0.090} \\
    \bottomrule
  \end{tabular}
  \label{tab:complementing}
  \end{scriptsize}
  \vspace{-0.3cm}
\end{table}

\textbf{Effect of complementing.} Table \ref{tab:complementing} shows an ablation study of the proposed complementing process to remove biased objects and prevent increasing FN of non-problematic classes (\emph{i.e.}, classes not including the biased problem). The first row is our baseline (\emph{i.e.}, RS+EPM \cite{jo2022recurseed}). Training a segmentation network with debiased labels improves at least 3.5\% of mIoU compared to our baseline RS+EPM \cite{jo2022recurseed} (rows 2 and 3). However, in row 2, the complementing process without the proposed WCE in \eqref{wce} significantly decreases FN but increases FP due to incorrect labels when complementing with the model's predictions. The last row achieves the best performance with considering certain masks, demonstrating the validity of the proposed components.

\begin{figure}
  \centering
  \includegraphics[width=0.95\linewidth]{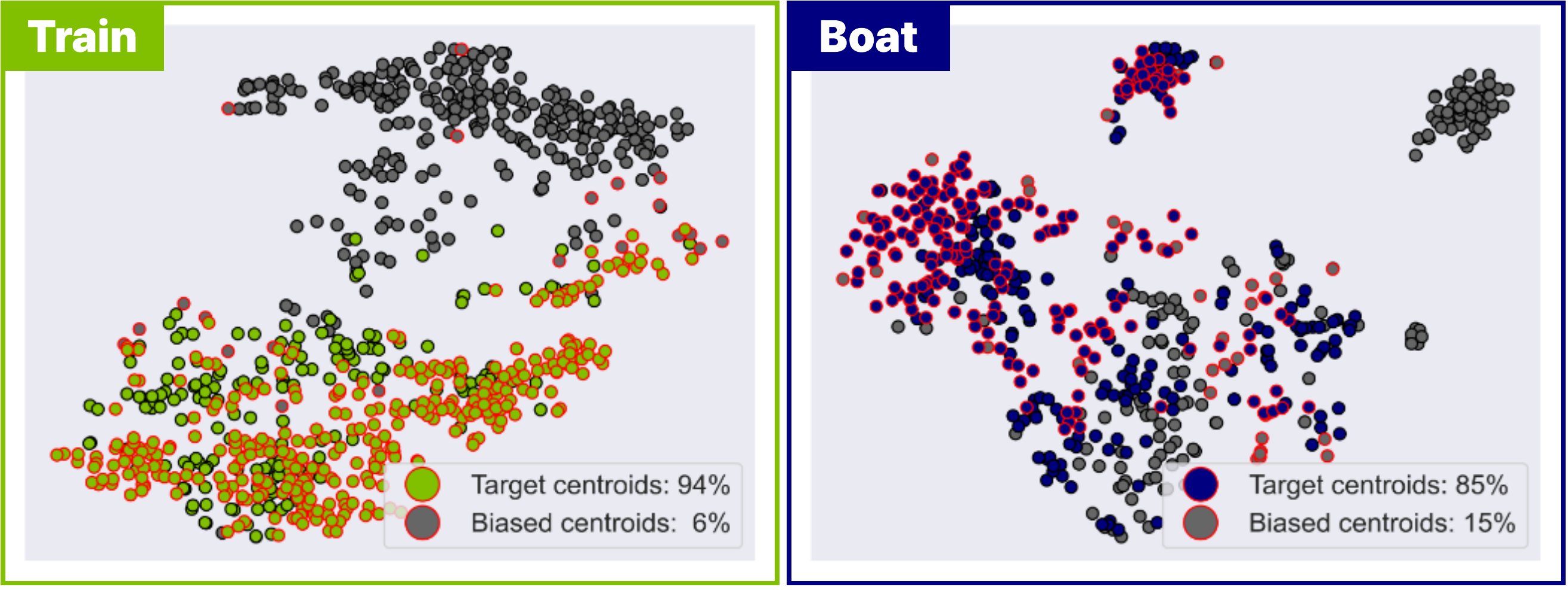}
  \caption{
      Visualization of selecting debiased centroids. We quantify the ratio of selecting target centroids by using pixel-wise annotations. The left and right results indicate train and boat classes, respectively. The percentage of target centroids is more than 85\%, proving the validity of the proposed selection.
  }
  \label{fig:tsne}
  \vspace{-0.4cm}
\end{figure}

\textbf{Reasoning of debiased centroids.} We quantify the ratio of target centroids in debiased centroids on the PASCAL VOC 2012 $train$ set. Fig. \ref{fig:bg_correlation} shows that K-means clustering separates two centroids (pink and orange) from decomposed embedding vectors for each class. We then measure each IoU score per centroid using pixel-wise annotations (each color has the IoU score). For simplicity, we classify all target and biased centroids based on their IoU scores, with target centroids having an IoU score above 0.3, biased centroids below 0.1, and others not visualized. Fig. \ref{fig:tsne} shows the visualization of target and biased centroids per class after dimensional reduction using T-SNE \cite{van2008visualizing}. The ratio of target centroids selected for all foreground classes is more than 85\% on the PASCAL VOC 2012 dataset (see other visualizations for all foreground classes in Appendix), validating the effectiveness of the proposed selection.

\begin{figure}
  \centering
  \includegraphics[width=1.0\linewidth, height=4cm]{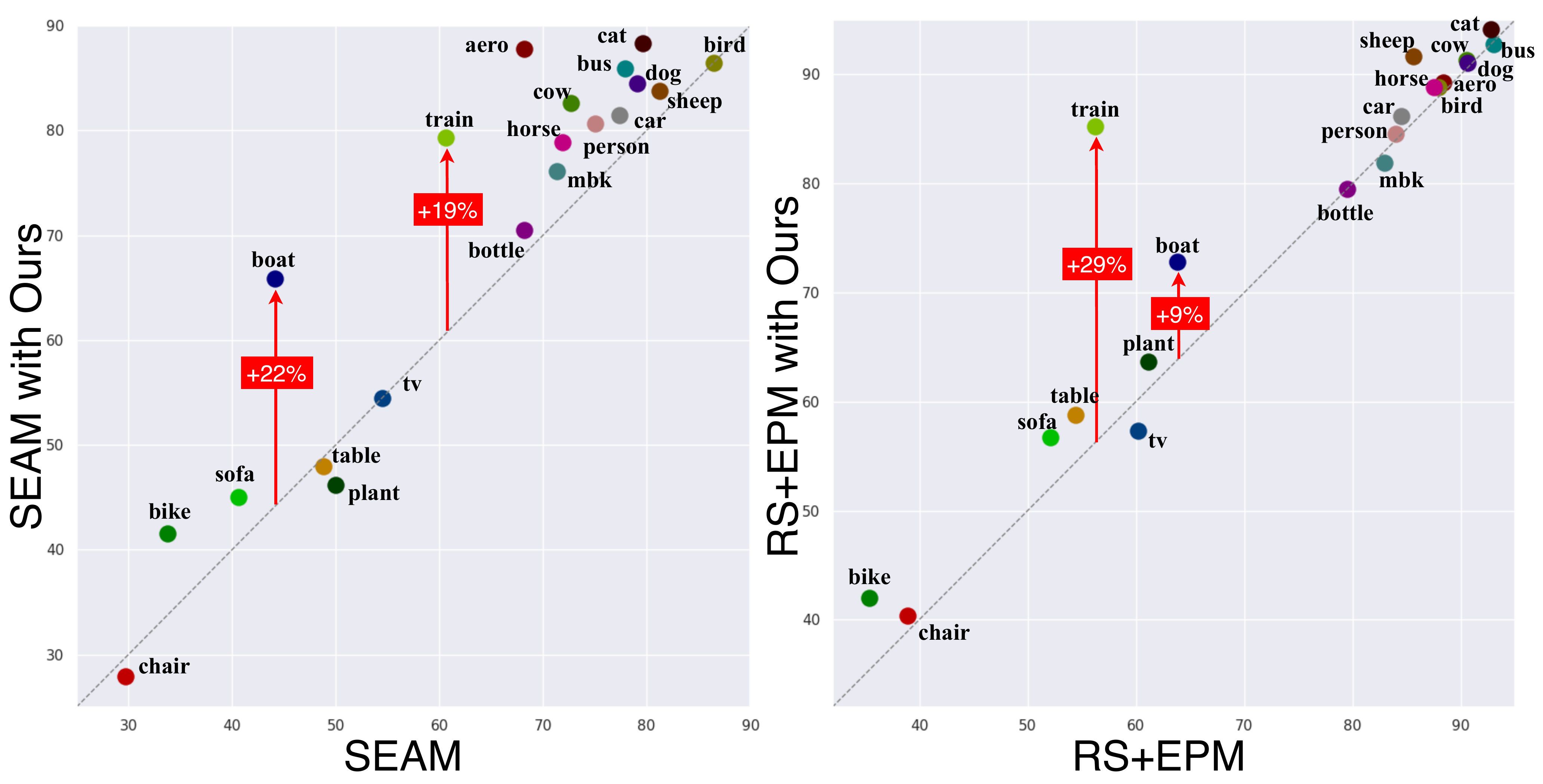}
  \caption{
      Category-wise comparison with SEAM \cite{wang2020self}, RS+EPM \cite{jo2022recurseed}, and ours in terms of the IoU (\%) on PASCAL VOC 2012 set. 
  }
  \label{fig:category}
  \vspace{-0.4cm}
\end{figure}

\textbf{Category-wise improvements.} Fig. \ref{fig:category} presents a class-wise comparison of our method with existing WSSS methods \cite{wang2020self, jo2022recurseed} on the PASCAL VOC 2012 validation set. Our method improves the mIoU scores of most categories. However, the performance of a few categories (\emph{e.g.}, tv/monitor) marginally decreases due to the poor quality of pseudo masks produced from the WSSS method. Notably, our method achieves significant improvements in the boat (+9\%) and train (+29\%) classes over RS+EPM \cite{jo2022recurseed}, demonstrating the superiority of our method in removing biased objects without additional supervision. We also provide class-wise improvements for other WSSS methods \cite{ahn2019weakly, lee2021anti} in Appendix.

\begin{figure}
  \centering
  \includegraphics[width=1.0\linewidth, height=4cm]{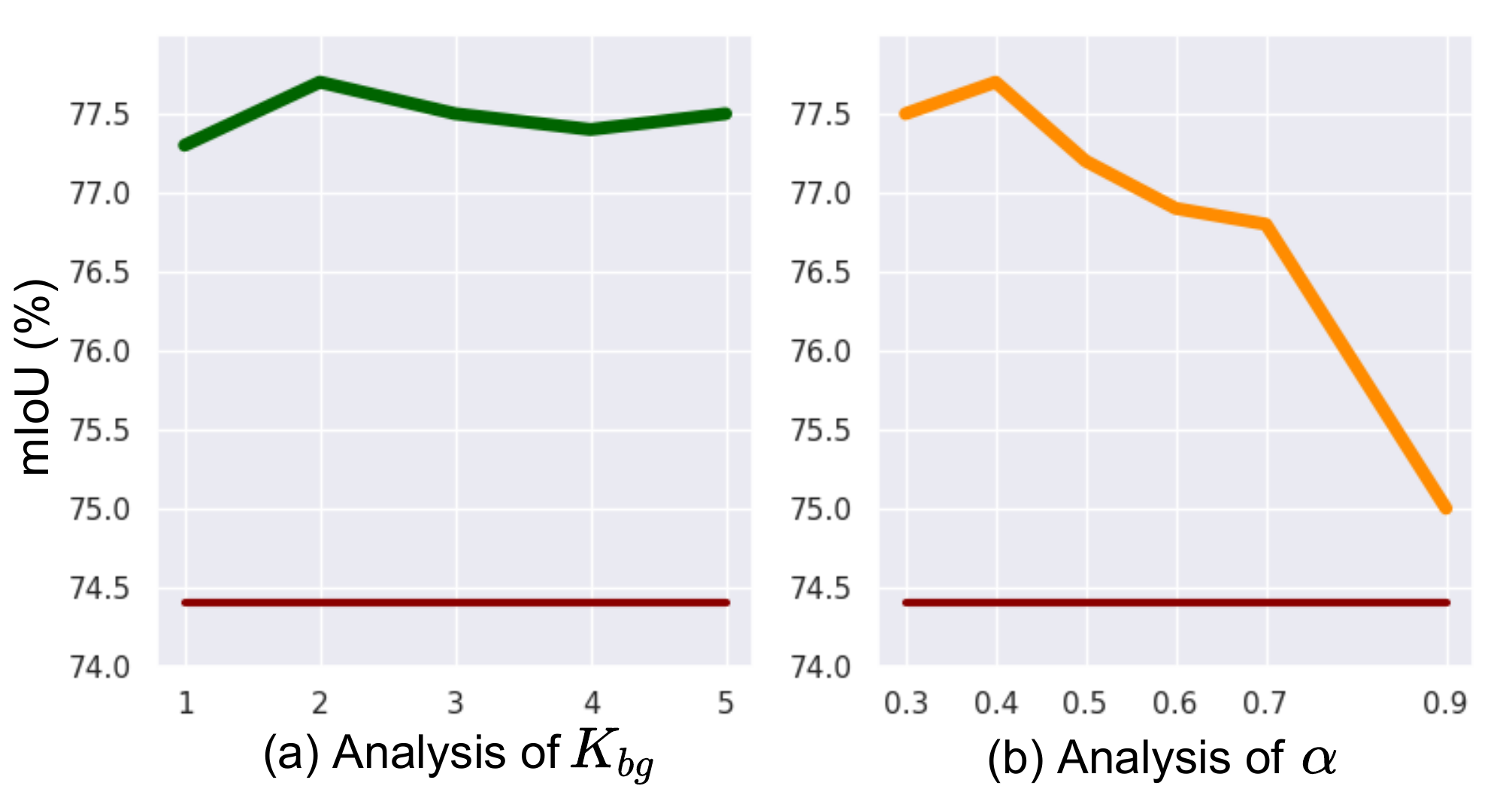}
  \caption{
      Sensitivity analysis of two hyperparameters $K_{bg}$ and $\alpha$. The mIoU scores are calculated on PASCAL VOC 2012 \emph{val} set. The red line is our baseline RS+EPM \cite{jo2022recurseed}.
  }
  \label{fig:hyper}
  \vspace{-0.4cm}
\end{figure}

\textbf{Hyperparameters.} We conduct the sensitivity analysis on two hyperparameters of our method, $K_{bg}$ and $\alpha$, using the PASCAL VOC 2012 validation set. Fig. \ref{fig:hyper} illustrates evaluation results. Our method improves performance across all hyperparameter settings compared to our baseline RS+EPM \cite{jo2022recurseed} (the red line). Varying $K_{bg}$ from 1 to 5 does not significantly affect our method's performance, indicating this hyperparameter's stability. On the other hand, larger values of $\alpha$ ($> 0.5$) result in only marginal improvements due to the difficulty in disentangling biased and target centroids. Conversely, smaller values of $\alpha$ ($< 0.5$) show sufficient improvements, demonstrating the validity of this hyperparameter to select debiased centroids based on the distance of all background centroids. These results further support the effectiveness of our method and provide insights for setting hyperparameters.


\section{Conclusion}\label{sec:concl}

In this work, we present MARS, a novel model-agnostic approach that addresses the biased problem in WSSS simply by exploiting the principle that USS-based information of biased objects can be easily matched with that of backgrounds of other samples. Accordingly, our approach significantly reduces FP due to WSSS bias, which is the primary reason that WSSS performance is limited compared to FSSS, achieves the fully-automatic biased removal without additional human resources, and complements debiased pixels with online predictions to avoid possible FN increases due to that removal. Thanks to following a model-agnostic manner, our approach yields consistent improvements when integrated with previous WSSS methods, narrowing the performance gap of 53\% between WSSS and FSSS. We believe the simplicity and effectiveness of our system will benefit future research in weakly-/semi-supervised tasks under the real industry with complex/multi-labels. 

\clearpage

{\small
\bibliographystyle{ieee_fullname}
\bibliography{main_brief}
}

\clearpage
\appendix

\begin{figure*}
  \centering
  \includegraphics[width=1.0\linewidth]{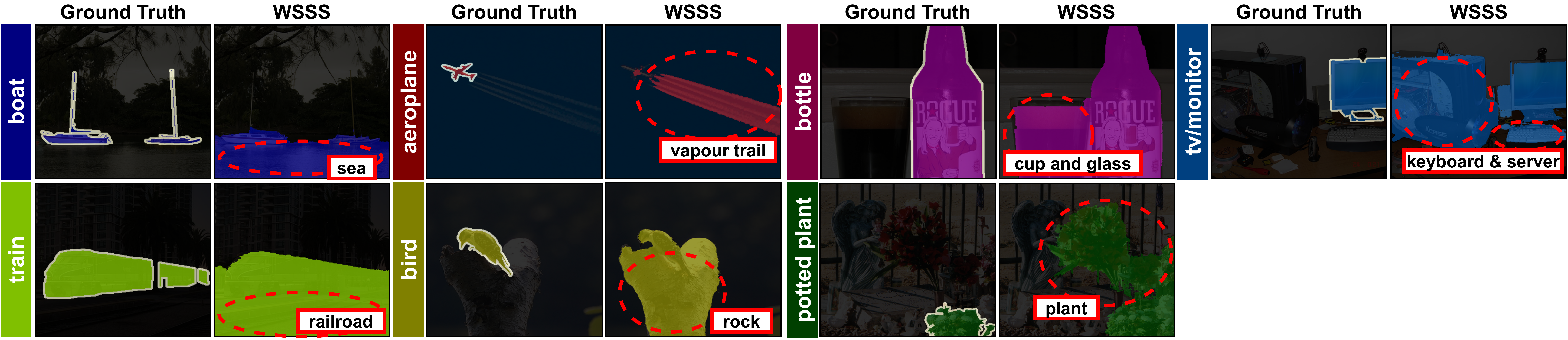}
  \caption{
      Examples of all biased objects on the PASCAL VOC 2012 dataset. Red dotted circles indicate the false activation of biased objects.
  }
  \label{fig:bias_examples}
\end{figure*}

\section{Additional Analysis}

\subsection{Examples of All Biased Objects}



In Fig. \ref{fig:problem}, we introduce two observations: (1) The severe FP of some classes causes the performance gap between existing WSSS methods \cite{wang2020self, jo2022recurseed} and FSSS, (2) 35\% of all classes (\emph{i.e.}, problematic classes) activate target objects (\emph{e.g.}, boat, train, bird, and aeroplane) with biased objects (\emph{e.g.}, sea, railroad, rock, and vapour trail). Following Fig. \ref{fig:problem}(c), we present additional examples of biased objects for all problematic classes in Fig. \ref{fig:bias_examples}. We hope that our detailed analysis of the biased problem in WSSS encourages other researchers to develop more robust and future WSSS approaches related to the biased problem.

\subsection{Effect of Selecting Debiased Centroids}



In Sec. \ref{sec:selecting}, we present selecting target objects among separated objects of all images after disentangling target and biased objects using the USS-based clustering in Sec. \ref{sec:selecting}. To evaluate the accuracy of debiased centroids, we measure how many selected centroids are target centroids among separated centroids of all images for each class in Fig. \ref{fig:additional_tsne}. Following Fig. \ref{fig:tsne} in Sec. \ref{sec:analysis}, we employ the T-SNE \cite{van2008visualizing} and the same criterion to classify target and biased centroids using pixel-wise annotations. In our experiments, the minimum accuracy for all classes on the PASCAL VOC 2012 $train$ dataset is 85\%. These results mean that the proposed selection using background information from other images successfully chooses target centroids in the group of target and biased centroids.

\begin{figure*}
  \centering
  \includegraphics[width=1.0\linewidth]{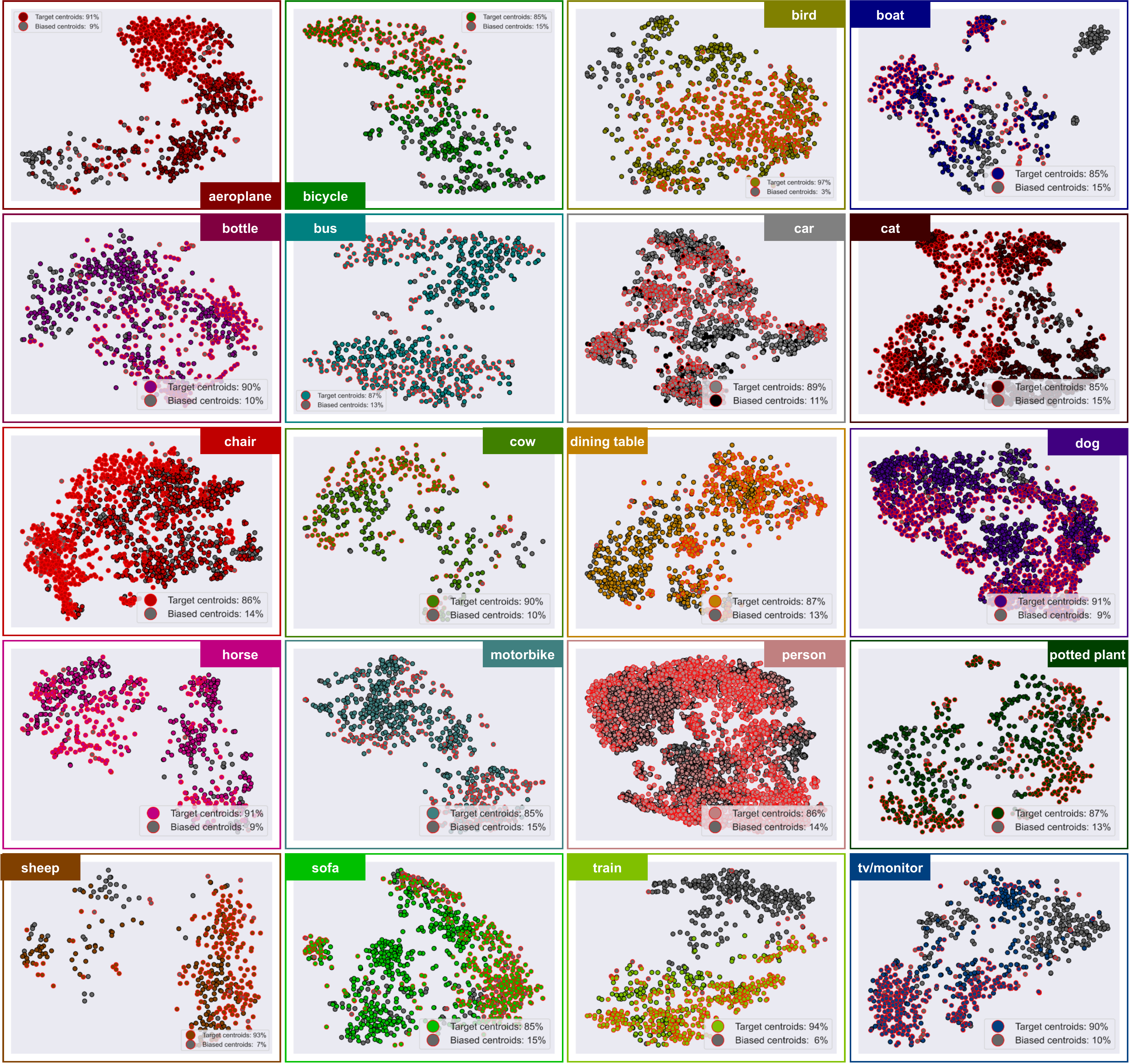}
  \caption{
      Visualization of selecting debiased centroids for all classes on the PASCAL VOC 2012 $train$ set. Red circles are selected centroids by our method. The average ratio of target centroids is more than 85\%, showing the effectiveness of the proposed selection.
  }
  \label{fig:additional_tsne}
\end{figure*}

\subsection{Additional Category-wise Improvements}

In line with Fig. \ref{fig:category}, we evaluate per-class improvements of four WSSS methods \cite{ahn2019weakly, wang2020self, lee2021anti, jo2022recurseed} with our method. All WSSS methods with ours show consistent improvements for top-3 classes (\emph{i.e.}, bicycle, train, and boat) in our FP analysis in Fig. \ref{fig:problem}(b). Also, the performance of non-problematic classes (\emph{e.g.}, person, dog, and cat) are improved by removing minor inconsistent objects (\emph{e.g.}, legs of the horse) when complementing debiased labels in Sec. \ref{sec:complementing}. However, a few categories (\emph{e.g.}, chair, dining table, and potted plant) show inconsistent improvements due to the poor quality of initial WSSS labels. As a result, our method improves less when the WSSS method performs erroneously, albeit our method improves performance for most categories.

\subsection{Qualitative Analysis with Existing Approaches}

\begin{figure*}
  \centering
  \includegraphics[width=0.95\linewidth]{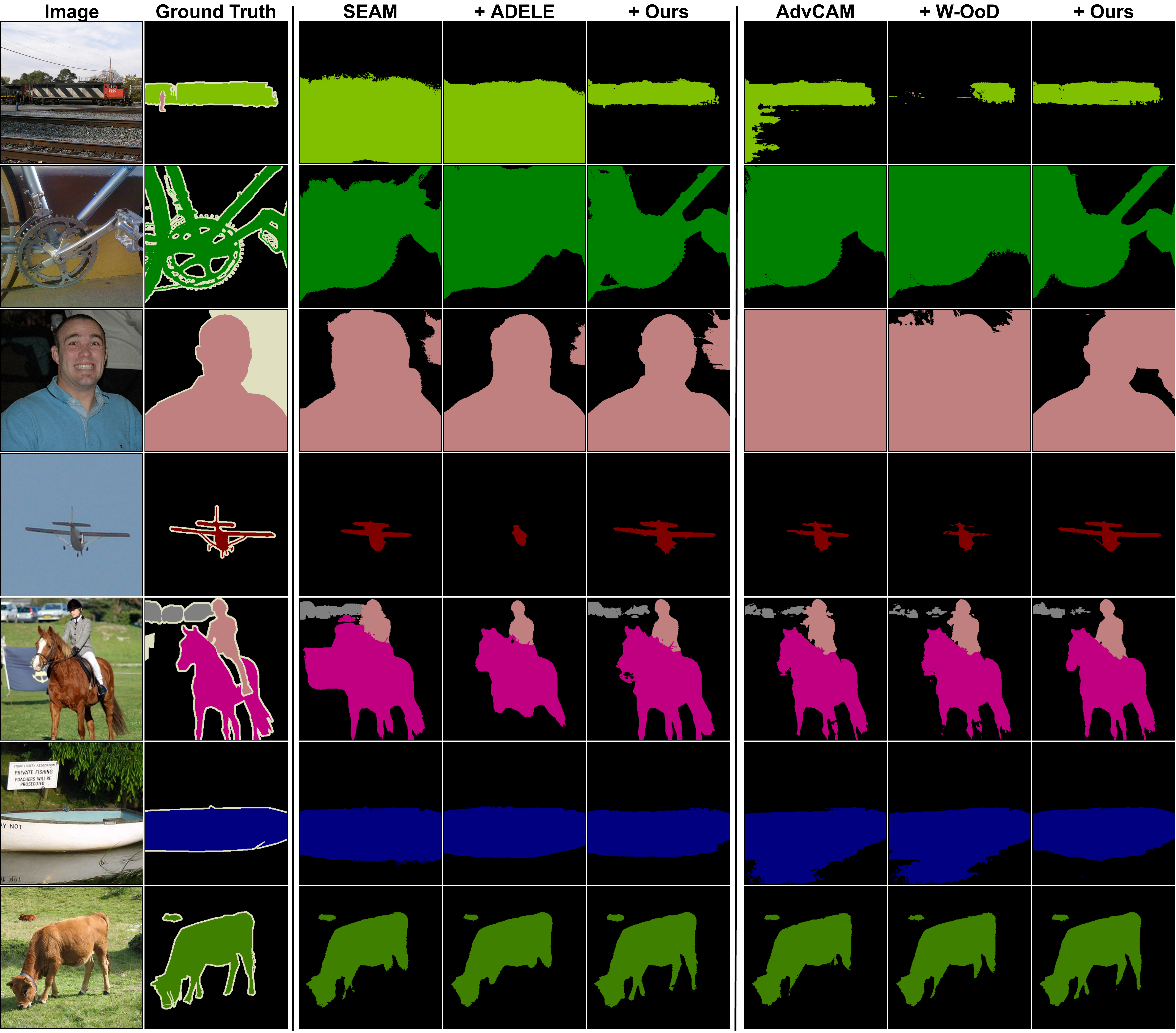}
  \caption{
      Examples of final segmentation results on PASCAL VOC 2012 \emph{val} set for SEAM \cite{wang2020self}, ADELE \cite{liu2022adaptive}, AdvCAM \cite{lee2021anti}, W-OoD \cite{lee2022weakly}, and Ours. 
  }
  \label{fig:comparison_wsss}
\end{figure*}

In addition to the quantitative comparison (see Table \ref{tab:wsss}), Fig. \ref{fig:comparison_wsss} illustrates a qualitative comparison of our method, ADELE \cite{liu2022adaptive}, and W-OoD \cite{lee2022weakly} using two WSSS methods \cite{wang2020self, lee2021anti}. ADELE \cite{liu2022adaptive} enlarges biased pixels since it enforces consistency of all classes without considering biased objects (the fourth column). Meanwhile, W-OoD \cite{lee2022weakly} removes biased objects (\emph{e.g.}, railroad) by utilizing extra images collected from human annotators, but it increases FN for most classes (\emph{e.g.}, train and aeroplane) due to implicitly training biased objects with collected images (the seventh column). Unlike these studies, to find biased pixels in WSSS labels, we first match biased objects with background information from other images by utilizing the USS features. Our MARS then complements biased pixels with the model's predictions to prevent increasing FN of non-biased pixels (\emph{e.g.}, legs of animals) in the fifth and eighth columns. Therefore, our method achieves the fully-automatic biased removal by explicitly eliminating biased objects in pseudo labels.

\begin{figure*}
  \centering
  \includegraphics[width=1.0\linewidth]{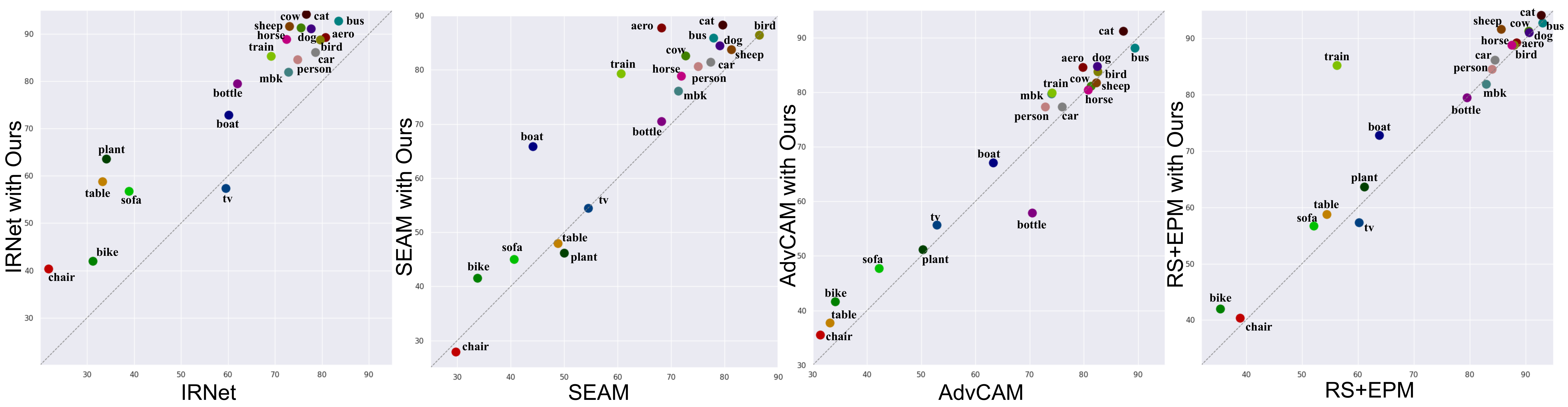}
  \caption{
      Category-wise comparison with IRNet \cite{ahn2019weakly}, SEAM \cite{wang2020self}, AdvCAM \cite{lee2021anti}, RS+EPM \cite{jo2022recurseed}, and ours in terms of the IoU (\%) on PASCAL VOC 2012 $train$ set. 
  }
  \label{fig:appendix_category}
  \vspace{-0.5cm}
\end{figure*}

\vspace{-0.2cm}

\section{Additional Results}

\subsection{Quantitative Results}



We present per-class segmentation results for two popular benchmarks in Tables \ref{tab:voc_val_detail}, \ref{tab:voc_test_detail}, and \ref{tab:coco_val_detail}. Our method significantly improves performance of train ($+29.1\%$) and boat ($+9.1\%$) classes, which suffer from the biased problem in Fig. \ref{fig:problem}, versus the previous state-of-the-art method (\emph{i.e.}, RS+EPM \cite{jo2022recurseed}). Also, we first demonstrate performance improvements for most classes including biased objects on the MS COCO 2014 dataset. When analyzing performance of our method on the MS COCO 2014 dataset, we find some classes (\emph{e.g.}, surfboard, tennis racket, and train) that contain biased objects (\emph{e.g.}, sea, tennis court, and railroad), causing performance degradation in existing WSSS methods \cite{huang2018weakly, jo2022recurseed}. By contrast, without additional human supervision, our method achieves significant improvements for most classes including surfboard ($+44.3\%$), tennis racket ($+43\%$), and train ($+24.6\%$) versus the latest WSSS method \cite{jo2022recurseed}. 


\subsection{Qualitative Results}

\begin{figure*}
  \centering
  \includegraphics[width=1.0\linewidth]{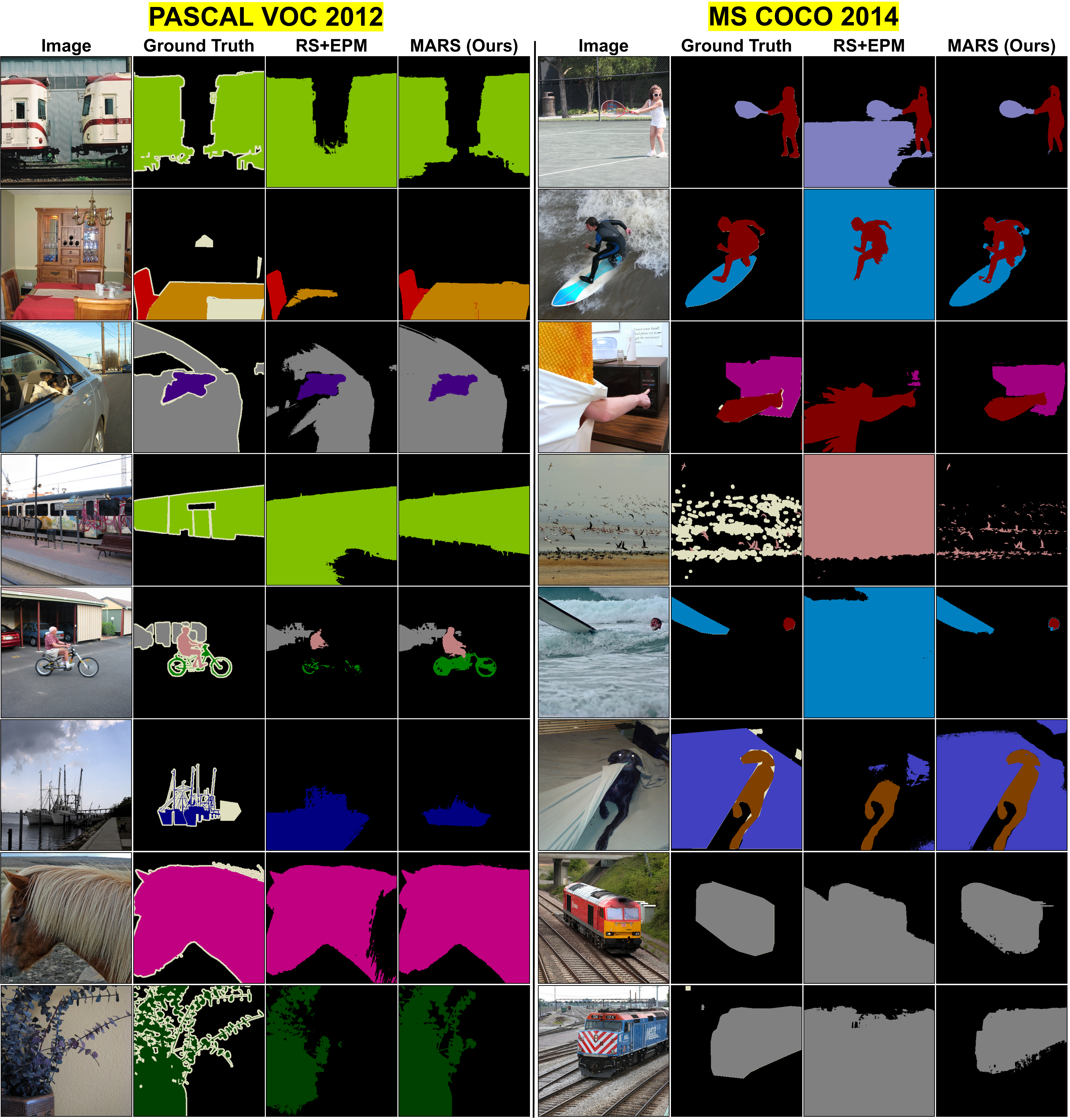}
  \caption{
      Qualitative segmentation results of the latest method (\emph{i.e.}, RS+EPM \cite{jo2022recurseed}) and the proposed MARS on PASCAL VOC 2012 and MS COCO 2014 validation sets.
  }
  \label{fig:final_wsss}
\end{figure*}

The qualitative segmentation results produced by the latest method \cite{jo2022recurseed} and our MARS are displayed in Fig. \ref{fig:final_wsss}. Our MARS performs well in various objects or multiple instances and can achieve satisfactory segmentation performance in challenging scenes. Specifically, our method removes biased objects for problematic classes (\emph{e.g.}, railroad in train, lake in boat, tennis court in tennis racket, and sea in surfboard), covers more object regions for large-scale objects (e.g., horse, car, and dining table), and captures the accurate boundaries of small-scale objects (e.g., bird) by complementing debiased labels with online predictions and considering the model's uncertainty. Our method shows superior performance in the qualitative and quantitative comparison with the previous state-of-the-art method (\emph{i.e.}, RS+EPM \cite{jo2022recurseed}), demonstrating the effectiveness of our MARS for the real-world dataset with multiple labels and complex relationships.

\clearpage

\begin{table*}[t]
  \centering
  \caption{
    Class-specific performance comparisons with WSSS methods in terms of IoUs (\%) on the PASCAL VOC 2012 \emph{val} set.
  }
  \begin{scriptsize}
  \begin{tabular}{
    p{0.15\textwidth} 
    p{0.015\textwidth} p{0.015\textwidth} p{0.015\textwidth} p{0.015\textwidth} p{0.015\textwidth} 
    p{0.015\textwidth} p{0.015\textwidth} p{0.015\textwidth} p{0.015\textwidth} p{0.015\textwidth} 
    p{0.015\textwidth} p{0.015\textwidth} p{0.015\textwidth} p{0.015\textwidth} p{0.015\textwidth} 
    p{0.015\textwidth} p{0.015\textwidth} p{0.015\textwidth} p{0.015\textwidth} p{0.015\textwidth} p{0.015\textwidth} 
    c }
    \toprule
    Method & \rotatebox[origin=l]{90}{bkg} & \rotatebox[origin=l]{90}{aero} & \rotatebox[origin=l]{90}{bike} & \rotatebox[origin=l]{90}{bird} & \rotatebox[origin=l]{90}{boat} & \rotatebox[origin=l]{90}{bottle} & \rotatebox[origin=l]{90}{bus} & \rotatebox[origin=l]{90}{car} & \rotatebox[origin=l]{90}{cat} & \rotatebox[origin=l]{90}{chair} & \rotatebox[origin=l]{90}{cow} & \rotatebox[origin=l]{90}{table} & \rotatebox[origin=l]{90}{dog} & \rotatebox[origin=l]{90}{horse} & \rotatebox[origin=l]{90}{mbk} & \rotatebox[origin=l]{90}{person} & \rotatebox[origin=l]{90}{plant} & \rotatebox[origin=l]{90}{sheep} & \rotatebox[origin=l]{90}{sofa} & \rotatebox[origin=l]{90}{train} & \rotatebox[origin=l]{90}{tv}  & mIoU  \\
    \hline \hline
    EM {\tiny ICCV'15} \cite{papandreou2015weakly} & 67.2 & 29.2 & 17.6 & 28.6 & 22.2 & 29.6 & 47.0 & 44.0 & 44.2 & 14.6 & 35.1 & 24.9 & 41.0 & 34.8 & 41.6 & 32.1 & 24.8 & 37.4 & 24.0 & 38.1 & 31.6 & 33.8 \\
    MIL-LSE {\tiny CVPR'15} \cite{pinheiro2015image} & 79.6 & 50.2 & 21.6 & 40.9 & 34.9 & 40.5 & 45.9 & 51.5 & 60.6 & 12.6 & 51.2 & 11.6 & 56.8 & 52.9 & 44.8 & 42.7 & 31.2 & 55.4 & 21.5 & 38.8 & 36.9 & 42.0 \\
    SEC {\tiny ECCV'16} \cite{kolesnikov2016seed} & 82.4 & 62.9 & 26.4 & 61.6 & 27.6 & 38.1 & 66.6 & 62.7 & 75.2 & 22.1 & 53.5 & 28.3 & 65.8 & 57.8 & 62.3 & 52.5 & 32.5 & 62.6 & 32.1 & 45.4 & 45.3 & 50.7 \\
    TransferNet {\tiny CVPR'16} \cite{hong2016learning} & 85.3 & 68.5 & 26.4 & 69.8 & 36.7 & 49.1 & 68.4 & 55.8 & 77.3 & 6.2 & 75.2 & 14.3 & 69.8 & 71.5 & 61.1 & 31.9 & 25.5 & 74.6 & 33.8 & 49.6 & 43.7 & 52.1 \\
    CRF-RNN {\tiny CVPR'17} \cite{roy2017combining} & 85.8 & 65.2 & 29.4 & 63.8 & 31.2 & 37.2 & 69.6 & 64.3 & 76.2 & 21.4 & 56.3 & 29.8 & 68.2 & 60.6 & 66.2 & 55.8 & 30.8 & 66.1 & 34.9 & 48.8 & 47.1 & 52.8 \\
    WebCrawl {\tiny CVPR'17} \cite{hong2017weakly} & 87.0 & 69.3 & 32.2 & 70.2 & 31.2 & 58.4 & 73.6 & 68.5 & 76.5 & 26.8 & 63.8 & 29.1 & 73.5 & 69.5 & 66.5 & 70.4 & 46.8 & 72.1 & 27.3 & 57.4 & 50.2 & 58.1 \\
    CIAN {\tiny AAAI'20} \cite{fan2020cian} & 88.2 & 79.5 & 32.6 & 75.7 & 56.8 & 72.1 & 85.3 & 72.9 & 81.7 & 27.6 & 73.3 & 39.8 & 76.4 & 77.0 & 74.9 & 66.8 & 46.6 & 81.0 & 29.1 & 60.4 & 53.3 & 64.3 \\
    SSDD {\tiny ICCV'19} \cite{shimoda2019self} & 89.0 & 62.5 & 28.9 & 83.7 & 52.9 & 59.5 & 77.6 & 73.7 & 87.0 & 34.0 & 83.7 & 47.6 & 84.1 & 77.0 & 73.9 & 69.6 & 29.8 & 84.0 & 43.2 & 68.0 & 53.4 & 64.9 \\
    PSA {\tiny CVPR'18} \cite{ahn2018learning} & 87.6 & 76.7 & 33.9 & 74.5 & 58.5 & 61.7 & 75.9 & 72.9 & 78.6 & 18.8 & 70.8 & 14.1 & 68.7 & 69.6 & 69.5 & 71.3 & 41.5 & 66.5 & 16.4 & 70.2 & 48.7 & 59.4 \\
    FickleNet {\tiny CVPR'19} \cite{lee2019ficklenet} & 89.5 & 76.6 & 32.6 & 74.6 & 51.5 & 71.1 & 83.4 & 74.4 & 83.6 & 24.1 & 73.4 & 47.4 & 78.2 & 74.0 & 68.8 & 73.2 & 47.8 & 79.9 & 37.0 & 57.3 & \textbf{64.6} & 64.9 \\
    RRM {\tiny AAAI'20} \cite{zhang2020reliability} & 87.9 & 75.9 & 31.7 & 78.3 & 54.6 & 62.2 & 80.5 & 73.7 & 71.2 & 30.5 & 67.4 & 40.9 & 71.8 & 66.2 & 70.3 & 72.6 & 49.0 & 70.7 & 38.4 & 62.7 & 58.4 & 62.6 \\
    SSSS {\tiny CVPR'20} \cite{araslanov2020single} & 88.7 & 70.4 & 35.1 & 75.7 & 51.9 & 65.8 & 71.9 & 64.2 & 81.1 & 30.8 & 73.3 & 28.1 & 81.6 & 69.1 & 62.6 & 74.8 & 48.6 & 71.0 & 40.1 & 68.5 & 64.3 & 62.7 \\
    SEAM {\tiny CVPR'20} \cite{wang2020self} & 88.8 & 68.5 & 33.3 & 85.7 & 40.4 & 67.3 & 78.9 & 76.3 & 81.9 & 29.1 & 75.5 & 48.1 & 79.9 & 73.8 & 71.4 & 75.2 & 48.9 & 79.8 & 40.9 & 58.2 & 53.0 & 64.5 \\
    AdvCAM {\tiny CVPR'21} \cite{lee2021anti} & 90.0 & 79.8 & 34.1 & 82.6 & 63.3 & 70.5 & 89.4 & 76.0 & 87.3 & 31.4 & 81.3 & 33.1 & 82.5 & 80.8 & 74.0 & 72.9 & 50.3 & 82.3 & 42.2 & 74.1 & 52.9 & 68.1 \\
    CPN {\tiny ICCV'21} \cite{zhang2021complementary} & 89.9 & 75.0 & 32.9 & 87.8 & 60.9 & 69.4 & 87.7 & 79.4 & 88.9 & 28.0 & 80.9 & 34.8 & 83.4 & 79.6 & 74.6 & 66.9 & 56.4 & 82.6 & 44.9 & 73.1 & 45.7 & 67.8 \\
    RIB {\tiny NeurIPS'21} \cite{lee2021reducing} & 90.3 & 76.2 & 33.7 & 82.5 & 64.9 & 73.1 & 88.4 & 78.6 & 88.7 & 32.3 & 80.1 & 37.5 & 83.6 & 79.7 & 75.8 & 71.8 & 47.5 & 84.3 & 44.6 & 65.9 & 54.9 & 68.3 \\
    AMN {\tiny CVPR'22} \cite{lee2022threshold} & 90.6 & 79.0 & 33.5 & 83.5 & 60.5 & 74.9 & 90.0 & 81.3 & 86.6 & 30.6 & 80.9 & 53.8 & 80.2 & 79.6 & 74.6 & 75.5 & 54.7 & 83.5 & 46.1 & 63.1 & 57.5 & 69.5 \\
    ADELE {\tiny CVPR'22} \cite{liu2022adaptive} & 91.1 & 77.6 & 33.0 & 88.9 & 67.1 & 71.7 & 88.8 & 82.5 & 89.0 & 26.6 & 83.8 & 44.6 & 84.4 & 77.8 & 74.8 & 78.5 & 43.8 & 84.8 & 44.6 & 56.1 & 65.3 & 69.3 \\
    W-OoD {\tiny CVPR'22} \cite{lee2022weakly} & 91.2 & 80.1 & 34.0 & 82.5 & 68.5 & 72.9 & 90.3 & 80.8 & 89.3 & 32.3 & 78.9 & 31.1 & 83.6 & 79.2 & 75.4 & 74.4 & 58.0 & 81.9 & 45.2 & 81.3 & 54.8 & 69.8 \\
    RCA {\tiny CVPR'22} \cite{zhou2022regional} & 91.8 & 88.4 & 39.1 & 85.1 & 69.0 & 75.7 & 86.6 & 82.3 & 89.1 & 28.1 & 81.9 & 37.9 & 85.9 & 79.4 & 82.1 & 78.6 & 47.7 & 84.4 & 34.9 & 75.4 & 58.6 & 70.6 \\
    SANCE {\tiny CVPR'22} \cite{li2022towards} & 91.4 & 78.4 & 33.0 & 87.6 & 61.9 & \textbf{79.6} & 90.6 & 82.0 & 92.4 & 33.3 & 76.9 & \textbf{59.7} & 86.4 & 78.0 & 76.9 & 77.7 & 61.1 & 79.4 & 47.5 & 62.1 & 53.3 & 70.9 \\
    MCTformer {\tiny CVPR'22} \cite{xu2022multi} & 91.9 & 78.3 & 39.5 & \textbf{89.9} & 55.9 & 76.7 & 81.8 & 79.0 & 90.7 & 32.6 & 87.1 & 57.2 & 87.0 & 84.6 & 77.4 & 79.2 & 55.1 & 89.2 & 47.2 & 70.4 & 58.8 & 71.9 \\
    RS+EPM {\tiny Arxiv'22} \cite{jo2022recurseed} & 92.2 & 88.4 & 35.4 & 87.9 & 63.8 & 79.5 & \textbf{93.0} & 84.5 & 92.7 & 39.0 & 90.5 & 54.5 & 90.6 & 87.5 & \textbf{83.0} & 84.0 & 61.1 & 85.6 & 52.1 & 56.2 & 60.2 & 74.4 \\
    \hline
    MARS (Ours) & \textbf{94.1} & \textbf{89.3} & \textbf{42.0} & 88.8 & \textbf{72.9} & 79.5 & 92.7 & \textbf{86.2} & \textbf{94.2} & \textbf{40.3} & \textbf{91.4} & 58.8 & \textbf{91.1} & \textbf{88.9} & 81.9 & \textbf{84.6} & \textbf{63.6} & \textbf{91.7} & \textbf{56.7} & \textbf{85.3} & 57.3 & \textbf{77.7} \\
    \hline
    \bottomrule
  \end{tabular}
  \end{scriptsize}
  \label{tab:voc_val_detail}
\end{table*}

\begin{table*}[t]
  \centering
  \caption{
    Class-specific performance comparisons with WSSS methods in terms of IoUs (\%) on the PASCAL VOC 2012 \emph{test} set.
  }
  \begin{scriptsize}
  \begin{tabular}{
    p{0.15\textwidth} 
    p{0.015\textwidth} p{0.015\textwidth} p{0.015\textwidth} p{0.015\textwidth} p{0.015\textwidth} 
    p{0.015\textwidth} p{0.015\textwidth} p{0.015\textwidth} p{0.015\textwidth} p{0.015\textwidth} 
    p{0.015\textwidth} p{0.015\textwidth} p{0.015\textwidth} p{0.015\textwidth} p{0.015\textwidth} 
    p{0.015\textwidth} p{0.015\textwidth} p{0.015\textwidth} p{0.015\textwidth} p{0.015\textwidth} p{0.015\textwidth} 
    c }
    \toprule
    Method & \rotatebox[origin=l]{90}{bkg} & \rotatebox[origin=l]{90}{aero} & \rotatebox[origin=l]{90}{bike} & \rotatebox[origin=l]{90}{bird} & \rotatebox[origin=l]{90}{boat} & \rotatebox[origin=l]{90}{bottle} & \rotatebox[origin=l]{90}{bus} & \rotatebox[origin=l]{90}{car} & \rotatebox[origin=l]{90}{cat} & \rotatebox[origin=l]{90}{chair} & \rotatebox[origin=l]{90}{cow} & \rotatebox[origin=l]{90}{table} & \rotatebox[origin=l]{90}{dog} & \rotatebox[origin=l]{90}{horse} & \rotatebox[origin=l]{90}{mbk} & \rotatebox[origin=l]{90}{person} & \rotatebox[origin=l]{90}{plant} & \rotatebox[origin=l]{90}{sheep} & \rotatebox[origin=l]{90}{sofa} & \rotatebox[origin=l]{90}{train} & \rotatebox[origin=l]{90}{tv}  & mIoU  \\
    \hline \hline
    EM {\tiny ICCV'15} \cite{papandreou2015weakly} & 76.3 & 37.1 & 21.9 & 41.6 & 26.1 & 38.5 & 50.8 & 44.9 & 48.9 & 16.7 & 40.8 & 29.4 & 47.1 & 45.8 & 54.8 & 28.2 & 30.0 & 44.0 & 29.2 & 34.3 & 46.0 & 39.6 \\
    MIL-LSE {\tiny CVPR'15} \cite{pinheiro2015image} & 78.7 & 48.0 & 21.2 & 31.1 & 28.4 & 35.1 & 51.4 & 55.5 & 52.8 & 7.8 & 56.2 & 19.9 & 53.8 & 50.3 & 40.0 & 38.6 & 27.8 & 51.8 & 24.7 & 33.3 & 46.3 & 40.6 \\
    SEC {\tiny ECCV'16} \cite{kolesnikov2016seed} & 83.5 & 56.4 & 28.5 & 64.1 & 23.6 & 46.5 & 70.6 & 58.5 & 71.3 & 23.2 & 54.0 & 28.0 & 68.1 & 62.1 & 70.0 & 55.0 & 38.4 & 58.0 & 39.9 & 38.4 & 48.3 & 51.7 \\
    TransferNet {\tiny CVPR'16} \cite{hong2016learning} & 85.7 & 70.1 & 27.8 & 73.7 & 37.3 & 44.8 & 71.4 & 53.8 & 73.0 & 6.7 & 62.9 & 12.4 & 68.4 & 73.7 & 65.9 & 27.9 & 23.5 & 72.3 & 38.9 & 45.9 & 39.2 & 51.2 \\
    CRF-RNN {\tiny CVPR'17} \cite{roy2017combining} & 85.7 & 58.8 & 30.5 & 67.6 & 24.7 & 44.7 & 74.8 & 61.8 & 73.7 & 22.9 & 57.4 & 27.5 & 71.3 & 64.8 & 72.4 & 57.3 & 37.3 & 60.4 & 42.8 & 42.2 & 50.6 & 53.7 \\
    WebCrawl {\tiny CVPR'17} \cite{hong2017weakly} & 87.2 & 63.9 & 32.8 & 72.4 & 26.7 & 64.0 & 72.1 & 70.5 & 77.8 & 23.9 & 63.6 & 32.1 & 77.2 & 75.3 & 76.2 & 71.5 & 45.0 & 68.8 & 35.5 & 46.2 & 49.3 & 58.7 \\
    PSA {\tiny CVPR'18} \cite{ahn2018learning} & 89.1 & 70.6 & 31.6 & 77.2 & 42.2 & 68.9 & 79.1 & 66.5 & 74.9 & 29.6 & 68.7 & 56.1 & 82.1 & 64.8 & 78.6 & 73.5 & 50.8 & 70.7 & 47.7 & 63.9 & 51.1 & 63.7 \\
    FickleNet {\tiny CVPR'19} \cite{lee2019ficklenet} & 90.3 & 77.0 & 35.2 & 76.0 & 54.2 & 64.3 & 76.6 & 76.1 & 80.2 & 25.7 & 68.6 & 50.2 & 74.6 & 71.8 & 78.3 & 69.5 & 53.8 & 76.5 & 41.8 & 70.0 & 54.2 & 65.0 \\
    SSDD {\tiny ICCV'19} \cite{shimoda2019self} & 89.5 & 71.8 & 31.4 & 79.3 & 47.3 & 64.2 & 79.9 & 74.6 & 84.9 & 30.8 & 73.5 & 58.2 & 82.7 & 73.4 & 76.4 & 69.9 & 37.4 & 80.5 & 54.5 & 65.7 & 50.3 & 65.5 \\
    RRM {\tiny AAAI'20} \cite{zhang2020reliability} & 87.8 & 77.5 & 30.8 & 71.7 & 36.0 & 64.2 & 75.3 & 70.4 & 81.7 & 29.3 & 70.4 & 52.0 & 78.6 & 73.8 & 74.4 & 72.1 & 54.2 & 75.2 & 50.6 & 42.0 & 52.5 & 62.9 \\
    SSSS {\tiny CVPR'20} \cite{araslanov2020single} & 88.7 & 70.4 & 35.1 & 75.7 & 51.9 & 65.8 & 71.9 & 64.2 & 81.1 & 30.8 & 73.3 & 28.1 & 81.6 & 69.1 & 62.6 & 74.8 & 48.6 & 71.0 & 40.1 & 68.5 & \textbf{64.3} & 62.7 \\
    SEAM {\tiny CVPR'20} \cite{wang2020self} & 88.8 & 68.5 & 33.3 & 85.7 & 40.4 & 67.3 & 78.9 & 76.3 & 81.9 & 29.1 & 75.5 & 48.1 & 79.9 & 73.8 & 71.4 & 75.2 & 48.9 & 79.8 & 40.9 & 58.2 & 53.0 & 64.5 \\
    AdvCAM {\tiny CVPR'21} \cite{lee2021anti} & 90.1 & 81.2 & 33.6 & 80.4 & 52.4 & 66.6 & 87.1 & 80.5 & 87.2 & 28.9 & 80.1 & 38.5 & 84.0 & 83.0 & 79.5 & 71.9 & 47.5 & 80.8 & 59.1 & 65.4 & 49.7 & 68.0 \\
    CPN {\tiny ICCV'21} \cite{zhang2021complementary} & 90.4 & 79.8 & 32.9 & 85.7 & 52.8 & 66.3 & 87.2 & 81.3 & 87.6 & 28.2 & 79.7 & 50.1 & 82.9 & 80.4 & 78.8 & 70.6 & 51.1 & 83.4 & 55.4 & 68.5 & 44.6 & 68.5 \\
    RIB {\tiny NeurIPS'21} \cite{lee2021reducing} & 90.4 & 80.5 & 32.8 & 84.9 & 59.4 & 69.3 & 87.2 & 83.5 & 88.3 & 31.1 & 80.4 & 44.0 & 84.4 & 82.3 & 80.9 & 70.7 & 43.5 & 84.9 & 55.9 & 59.0 & 47.3 & 68.6 \\
    AMN {\tiny CVPR'22} \cite{lee2022threshold} & 90.7 & 82.8 & 32.4 & 84.8 & 59.4 & 70.0 & 86.7 & 83.0 & 86.9 & 30.1 & 79.2 & 56.6 & 83.0 & 81.9 & 78.3 & 72.7 & 52.9 & 81.4 & 59.8 & 53.1 & 56.4 & 69.6 \\
    W-OoD {\tiny CVPR'22} \cite{lee2022weakly} & 91.4 & 85.3 & 32.8 & 79.8 & 59.0 & 68.4 & 88.1 & 82.2 & 88.3 & 27.4 & 76.7 & 38.7 & 84.3 & 81.1 & 80.3 & 72.8 & 57.8 & 82.4 & 59.5 & \textbf{79.5} & 52.6 & 69.9 \\
    RCA {\tiny CVPR'22} \cite{zhou2022regional} & 92.1 & 86.6 & 40.0 & 90.1 & 60.4 & 68.2 & 89.8 & 82.3 & 87.0 & 27.2 & 86.4 & 32.0 & 85.3 & 88.1 & 83.2 & 78.0 & 59.2 & 86.7 & 45.0 & 71.3 & 52.5 & 71.0 \\
    SANCE {\tiny CVPR'22} \cite{li2022towards} & 91.6 & 82.6 & 33.6 & 89.1 & 60.6 & \textbf{76.0} & 91.8 & 83.0 & 90.9 & 33.5 & 80.2 & 64.7 & 87.1 & 82.3 & 81.7 & 78.3 & 58.5 & 82.9 & \textbf{60.9} & 53.9 & 53.5 & 72.2 \\
    MCTformer {\tiny CVPR'22} \cite{xu2022multi} & 92.3 & 84.4 & 37.2 & 82.8 & 60.0 & 72.8 & 78.0 & 79.0 & 89.4 & 31.7 & 84.5 & 59.1 & 85.3 & 83.8 & 79.2 & 81.0 & 53.9 & 85.3 & 60.5 & 65.7 & 57.7 & 71.6 \\
    RS+EPM {\tiny Arxiv'22} \cite{jo2022recurseed} & 91.9 & 89.7 & 37.3 & 88.0 & 62.5 & 72.1 & 93.5 & 85.6 & 90.2 & 36.3 & \textbf{88.3} & 62.5 & 86.3 & \textbf{89.1} & 82.9 & 81.2 & 59.7 & 89.2 & 56.2 & 44.5 & 59.4 & 73.6 \\
    \hline
    MARS (Ours) & \textbf{93.7} & \textbf{93.3} & \textbf{40.3} & \textbf{90.8} & \textbf{70.8} & 71.7 & \textbf{94.0} & \textbf{86.3} & \textbf{93.9} & \textbf{40.4} & 87.6 & \textbf{67.6} & \textbf{90.0} & 87.3 & \textbf{83.9} & \textbf{83.1} & \textbf{64.2} & \textbf{89.5} & 59.6 & 79.0 & 55.1 & \textbf{77.2} \\
    \hline
    \bottomrule
  \end{tabular}
  \end{scriptsize}
  \label{tab:voc_test_detail}
\end{table*}

\clearpage

\begin{table*}[t]
  \centering
  \caption{
    Class-specific performance comparisons with WSSS methods in terms of IoUs (\%) on the MS COCO 2014 \emph{val} set.
  }
  \begin{scriptsize}
  \begin{tabular}{
    p{0.120\textwidth} c c c c | p{0.120\textwidth} c c c c}
    \toprule
    Class & SEC \cite{kolesnikov2016seed} & DSRG \cite{huang2018weakly} & RS+EPM \cite{jo2022recurseed} & MARS (Ours) & Class & SEC \cite{kolesnikov2016seed} & DSRG \cite{huang2018weakly} & RS+EPM \cite{jo2022recurseed} & MARS (Ours) \\
    \hline \hline
    background & 74.3 & 80.6 & 83.6 & \textbf{83.7} & wine glass & 22.3 & 24.0 & 39.8 & \textbf{45.5} \\
    person & 43.6 & - & \textbf{74.9} & 56.8 & cup & 17.9 & 20.4 & 38.9 & \textbf{42.0} \\
    bicycle & 24.2 & 30.4 & 55.0 & \textbf{59.2} & fork & 1.8 & 0.0 & \textbf{4.9} & 1.7 \\
    car & 15.9 & 22.1 & 50.1 & \textbf{52.0} & knife & 1.4 & 5.0 & \textbf{9.0} & 6.4 \\
    motorcycle & 52.1 & 54.2 & 72.9 & \textbf{75.2} & spoon & 0.6 & 0.5 & \textbf{1.1} & 0.9 \\
    airplane & 36.6 & 45.2 & 76.5 & \textbf{79.6} & bowl & 12.5 & \textbf{18.8} & 11.3 & 14.1 \\
    bus & 37.7 & 38.7 & 72.5 & \textbf{76.8} & banana & 43.6 & 46.4 & 67.0 & \textbf{67.7} \\
    train & 30.1 & 33.2 & 47.4 & \textbf{72.0} & apple & 23.6 & 24.3 & \textbf{49.2} & 47.9 \\
    truck & 24.1 & 25.9 & 46.5 & \textbf{54.1} & sandwich & 22.8 & 24.5 & 33.7 & \textbf{34.9} \\
    boat & 17.3 & 20.6 & 44.1 & \textbf{52.1} & orange & 44.3 & 41.2 & 62.3 & \textbf{62.5} \\
    traffic light & 16.7 & 16.1 & \textbf{60.8} & 53.8 & broccoli & 36.8 & 35.7 & \textbf{50.4} & 45.9 \\
    fire hydrant & 55.9 & 60.4 & 80.3 & \textbf{80.9} & carrot & 6.7 & 15.3 & \textbf{35.0} & 31.7 \\
    stop sign & 48.4 & 51.0 & \textbf{84.1} & 76.8 & hot dog & 31.2 & 24.9 & 48.3 & \textbf{51.5} \\
    parking meter & 25.2 & 26.3 & \textbf{77.8} & 74.8 & pizza & 50.9 & 56.2 & \textbf{68.6} & 68.0 \\
    bench & 16.4 & 22.3 & 41.2 & \textbf{47.2} & donut & 32.8 & 34.2 & 62.3 & \textbf{64.9} \\
    bird & 34.7 & 41.5 & 62.6 & \textbf{72.3} & cake & 12.0 & 6.9 & 48.3 & \textbf{53.3} \\
    cat & 57.2 & 62.2 & 79.2 & \textbf{80.9} & chair & 7.8 & 9.7 & 28.9 & \textbf{30.3} \\
    dog & 45.2 & 55.6 & 73.3 & \textbf{76.3} & couch & 5.6 & 17.7 & 44.9 & \textbf{49.1} \\
    horse & 34.4 & 42.3 & 76.1 & \textbf{78.2} & potted plant & 6.2 & 14.3 & 16.9 & \textbf{20.6} \\
    sheep & 40.3 & 47.1 & 80.0 & \textbf{83.5} & bed & 23.4 & 32.4 & 53.6 & \textbf{55.9} \\
    cow & 41.4 & 49.3 & 79.3 & \textbf{83.2} & dining table & 0.0 & 3.8 & \textbf{24.6} & 17.4 \\
    elephant & 62.9 & 67.1 & 85.6 & \textbf{87.7} & toilet & 38.5 & 43.6 & 71.1 & \textbf{76.5} \\
    bear & 59.1 & 62.6 & 82.9 & \textbf{87.5} & tv & 19.2 & 25.3 & 49.9 & \textbf{54.9} \\
    zebra & 59.8 & 63.2 & 87.0 & \textbf{87.9} & laptop & 20.1 & 21.1 & 56.6 & \textbf{64.5} \\
    giraffe & 48.8 & 54.3 & 82.2 & \textbf{83.4} & mouse & 3.5 & 0.9 & \textbf{17.4} & 12.9 \\
    backpack & 0.3 & 0.2 & 9.4 & \textbf{11.9} & remote & 17.5 & 20.6 & 54.8 & \textbf{55.3} \\
    umbrella & 26.0 & 35.3 & 73.4 & \textbf{77.1} & keyboard & 12.5 & 12.3 & 48.8 & \textbf{51.8} \\
    handbag & 0.5 & 0.7 & 4.6 & \textbf{8.4} & cell phone & 32.1 & 33.0 & 60.8 & \textbf{64.6} \\
    tie & 6.5 & 7.0 & 17.2 & \textbf{18.4} & microwave & 8.2 & 11.2 & 43.6 & \textbf{56.9} \\
    suitcase & 16.7 & 23.4 & 53.9 & \textbf{57.2} & oven & 13.7 & 12.4 & 38.0 & \textbf{43.5} \\
    frisbee & 12.3 & 13.0 & \textbf{57.7} & 57.5 & toaster & 0.0 & 0.0 & 0.0 & 0.0 \\
    skis & 1.6 & 1.5 & 8.2 & \textbf{10.8} & sink & 10.8 & 17.8 & 36.9 & \textbf{40.7} \\
    snowboard & 5.3 & 16.3 & 24.7 & \textbf{27.7} & refrigerator & 4.0 & 15.5 & 51.8 & \textbf{63.4} \\
    sports ball & 7.9 & 9.8 & \textbf{41.6} & 40.4 & book & 0.4 & 12.3 & 27.3 & \textbf{29.2} \\
    kite & 9.1 & 17.4 & 62.6 & \textbf{63.8} & clock & 17.8 & 20.7 & \textbf{23.3} & 19.8 \\
    baseball bat & 1.0 & \textbf{4.8} & 1.5 & 1.6 & vase & 18.4 & 23.9 & 26.0 & \textbf{31.0} \\
    baseball glove & 0.6 & \textbf{1.2} & 0.4 & 0.3 & scissors & 16.5 & 17.3 & \textbf{47.1} & 47.0 \\
    skateboard & 7.1 & 14.4 & 34.8 & \textbf{34.9} & teddy bear & 47.0 & 46.3 & 68.8 & \textbf{69.5} \\
    surfboard & 7.7 & 13.5 & 17.0 & \textbf{61.3} & hair drier & 0.0 & 0.0 & 0.0 & 0.0 \\
    tennis racket & 9.1 & 6.8 & 9.0 & \textbf{52.0} & toothbrush & 2.8 & 2.0 & 19.7 & \textbf{32.2} \\\cline{6-10}
    bottle & 13.2 & 22.3 & \textbf{38.1} & 36.6 & \textbf{mIoU} & 22.4 & 26.0 & 46.4 & \textbf{49.4} \\
    
    \bottomrule
  \end{tabular}
  \end{scriptsize}
  \label{tab:coco_val_detail}
\end{table*}

\clearpage

\end{document}